\documentclass[lettersize,journal]{IEEEtran}
\usepackage{amsmath,amsfonts}
\usepackage{algorithmic}
\usepackage{algorithm}
\usepackage{array}
\usepackage[caption=false,font=normalsize,labelfont=sf,textfont=sf]{subfig}
\usepackage{textcomp}
\usepackage{stfloats}
\usepackage{url}
\usepackage{verbatim}
\usepackage{graphicx}
\usepackage{cite}
\usepackage{framed,multirow}
\usepackage{booktabs}
\usepackage{color}
\graphicspath{{./figures/}}
\begin{document}

\title{Learning Feature Matching via Matchable Keypoint-Assisted Graph Neural Network}

\author{Zizhuo Li and Jiayi Ma, \emph{Senior Member, IEEE}}

\markboth{}%
{Shell \MakeLowercase{\textit{et al.}}: A Sample Article Using IEEEtran.cls for IEEE Journals}


\maketitle

\begin{abstract}

Accurately matching local features between a pair of images corresponding to the same 3D scene is a challenging computer vision task. Previous studies typically utilize attention-based graph neural networks (GNNs) with fully-connected graphs over keypoints within/across images for visual and geometric information reasoning. However, in the context of feature matching, a significant number of keypoints are non-repeatable due to occlusion and failure of the detector, and thus irrelevant for message passing. The connectivity with non-repeatable keypoints not only introduces redundancy, resulting in limited efficiency (quadratic computational complexity \emph{w.r.t.} the keypoint number), but also interferes with the representation aggregation process, leading to limited accuracy. Targeting towards high accuracy and efficiency, we propose MaKeGNN, a sparse attention-based GNN architecture which bypasses non-repeatable keypoints and leverages matchable ones to guide compact and meaningful message passing. More specifically, our \emph{Bilateral Context-Aware Sampling} (BCAS) \emph{Module} first dynamically samples two small sets of well-distributed keypoints with high matchability scores from the image pair. Then, our \emph{Matchable Keypoint-Assisted Context Aggregation} (MKACA) \emph{Module} regards sampled informative keypoints as message bottlenecks and thus constrains each keypoint only to retrieve favorable contextual information from intra- and inter- matchable keypoints, evading the interference of irrelevant and redundant connectivity with non-repeatable ones. Furthermore, considering the potential noise in initial keypoints and sampled matchable ones, the MKACA module adopts a matchability-guided attentional aggregation operation for purer data-dependent context propagation. By these means, we achieve the state-of-the-art performance on relative camera estimation, fundamental matrix estimation, and visual localization, while significantly reducing computational and memory complexity compared to typical attentional GNNs.

\end{abstract}


\section{Introduction}\label{introduction}
\IEEEPARstart{F}{eature} matching which refers to establishing reliable correspondences between images depicting the same visual content, continues to be a cornerstone for a board range of downstream vision applications~\cite{ma2021image}, such as Simultaneous
Localization and Mapping (SLAM)~\cite{mur2015orb}, Structure-from-Motion (SfM)~\cite{schonberger2016structure}, visual localization~\cite{sarlin2019coarse}, and image fusion~\cite{zhang2021image}. Given a pair of images, the most standard pipeline starts with detecting keypoints and constructing local descriptors, so that point-to-point correspondences can be generated by searching the nearest neighbor (NN) in line with the similarity of feature descriptors. However, due to visual ambiguity caused by poor texture, repetitive elements, viewpoint change, illumination variation, and motion blur, the matches given by an NN matcher are inevitably dominated by false ones (\emph{i.e.}, outliers), resulting in unpromising accuracy~\cite{sarlin2020superglue,lu2023robust}. To alleviate this issue, a crowd of approaches recently focus on designing learnable outlier rejection strategies~\cite{yi2018learning,zhang2019learning,sun2020acne,zhao2021progressive,liu2021learnable,dai2022ms2dg,zheng2022msa,zhang2023convmatch, li2023umatch,li2023two,liu2023pgfnet} to filter spurious matches from candidate ones. Specifically, most of them adopt a PointNet-like~\cite{qi2017pointnet} architecture, \emph{i.e.}, multi-layer perceptrons (MLPs), to process each correspondence individually and infer the probability of correspondences to be correct ones (\emph{i.e.}, inliers). Despite the inspiring progress, these methods are inherently limited by the following two aspects: 1) Their inputs are putative correspondences provided by an NN search, so it is impossible to retrieve correct matches beyond these ones, which means that the upper bound on the performance is limited by the quality of initial matches; 2) Local visual descriptors and correspondence coordinates are separately used for match generation and mismatch rejection, respectively, neglecting the interaction between visual information from high-level image representation and geometric information from 2D keypoint distribution, calling for the research towards the design of effective feature matching methods to break the limitation of vanilla nearest neighbor correspondences. To this end, a line of research directly learns to find the partial assignment between two sets of local features with an attention-based Graph Neural Network (GNN)~\cite{suwanwimolkul2022efficient,cai2023htmatch}. Representatively, SuperGlue~\cite{sarlin2020superglue} constructs densely-connected graphs over intra- and inter-image keypoints and performs message passing by means of self- and cross-attention. Since SuperGlue encodes both spatial relationships of keypoints and their associated visual appearance, it essentially produces more discriminative representations for feature matching. However, the remarkable performance of SuperGlue comes with high computational complexity of $\mathcal{O}(N^2)$ for matrix multiplications and memory occupation of $\mathcal{O}(N^2)$ to save the attention weights, where $N$ denotes the number of nodes, substantially degrading its efficiency in real-time applications and making it unsuitable for mobile deployment. Additionally, a large number of detected keypoints are non-repeatable and thus do not have correspondences in the other image (about 70\% of 2$k$ keypoints in YFCC100M dataset~\cite{thomee2016yfcc100m}) due to occlusion, viewpoint change, and failure of the detector, \emph{etc.}, which often occur in feature matching. This means that such fully-connected graphs actually introduce too many redundant and meaningless connections, which to some degree limit the networks' capability for fine-grained feature update. To mitigate these downsides, a plethora of follow-ups, such as SGMNet~\cite{chen2021learning} and ClusterGNN~\cite{shi2022clustergnn}, have made notable contributions to the design of more efficient and compact context aggregation operations by relaxing densely-connected graphs into sparsely-connected ones, but they are significantly less accurate. In particular, SGMNet initially generates a set of seed matches using an NN matcher, which are unfortunately contaminated by heavy outliers, making the aggregated information not pure. Although a weighted attentional aggregation is further introduced to alleviate this problem, it subsequently results in massive loss of contextual cues during the unpooling process. Also, ClusterGNN conducts cluster-wise matching using a learnable coarse-to-fine paradigm. However, the generated clusters lack clear geometric meanings, and the clustering process does not exclude non-repeatable keypoints.

In the context of feature matching, a keypoint aggregates contextual information from nodes connected to it through an attentional GNN. However, as stated above, the connectivity with non-repeatable keypoints is unnecessary and distracting, which not only reduces efficiency but also degrades accuracy. \emph{Thus, it is natural to ask that: can a keypoint only exchange messages with repeatable and informative keypoints (matchable keypoints\footnote{A keypoint that exists in co-visible areas and has a correspondence in the other image, is considered matchable. Matchable keypoints can provide fruitful geometric and visual cues, which are of great benefit for feature matching.}) to achieve fine-grained feature update?}

Following this train of thought, we propose a \textbf{Ma}tchable \textbf{Ke}ypoint-Assisted \textbf{G}raph \textbf{N}eural \textbf{N}etwork (MaKeGNN) that accounts for both accuracy and efficiency to learning feature matching. First, we introduce a \emph{Bilateral Context-Aware Sampling} (BCAS) \emph{Module} to dynamically sample two set of well-distributed matchable keypoints from a pair of images, as shown in Figs.~\ref{framework} and \ref{sampler}. Second, we design a \emph{Matchable Keypoint-Assisted Context Aggregation} (MKACA) \emph{Module} whose graph structure is greatly sparse to achieve efficient and effective message passing. In particular, rather than conducting communication with all features within and across images, the MKACA module considers the selected matchable keypoints as bottlenecks and constrains each keypoint only attend to intra- and inter- matchable ones, avoiding the interference of irrelevant connections, relieving the computational burden, and retrieving compact and fine-grained context. Meanwhile, we notice that the potential noise in initial keypoints and sampled matchable ones may lead to impure information broadcast and thus integrate a matchability-guided attentional aggregation mechanism into the MKACA module for cleaner and sharper data-dependent message propagation. Importantly, with the proposed BCAS and MKACA modules, the computational complexity of attention is significantly reduced from $\mathcal{O}(N^2)$ to $\mathcal{O}(N\textit{k})$, where \textit{k} is the number of sampled matchable keypoints from each image. Comprehensive experiments on
multiple visual tasks such as relative camera estimation, fundamental matrix estimation and visual localization reveal that our method
outperforms long lines of prior work.

We summarize our contributions as follows:
\begin{itemize}
	\item We propose a \emph{Bilateral Context-Aware Sampling} (BCAS) \emph{Module}, which enables the model to dynamically select well-distributed keypoints with high matchability scores, to guide compact and robust context aggregation.
	\item We introduce a \emph{Matchable Keypoint-Assisted Context Aggregation} (MKACA) \emph{Module} for message passing. Instead of constructing fully-connected graphs, we leverage the sampled informative keypoints as bottlenecks, so that each keypoint only communicates with intra- and inter- matchable ones to update per-keypoint feature elegantly, which not only increases efficiency but also accuracy.
	\item Our MakeGNN shows superior performance over the state-of-the-arts on relative camera estimation, fundamental matrix estimation, and visual localization.
\end{itemize}


\section{Previous Arts}\label{Previous Arts}
\subsection{Local Feature Matching}\label{Local_Feature_Matching}
Matching local features between a pair of images is key to various applications including relative pose estimation and visual localization. Traditionally, keypoints are first detected from each image and then patches centered at these ones are used to generate corresponding visual descriptors with hand-crafted methods such as SIFT~\cite{lowe2004distinctive} and ORB~\cite{rublee2011orb}. Finally, an NN matcher and its variants, \emph{e.g.}, mutual nearest neighbor (MNN) check and ratio test (RT)~\cite{lowe2004distinctive} are used to establish correspondences between two sets of keypoints. With the advent of the deep learning era, the robustness of local features under illumination variation and viewpoint change can be significantly improved by learning-based methods, such as SuperPoint~\cite{detone2018superpoint}, D2-Net~\cite{dusmanu2019d2}, R2D2~\cite{revaud2019r2d2}, and AWDesc~\cite{wang2023attention}. In addition to learning-based detector-descriptors, some works recently focus on learning local feature matching~\cite{sarlin2020superglue,chen2021learning,suwanwimolkul2022efficient,shi2022clustergnn,cai2023htmatch,lu2023paraformer}. As a pioneering work, SuperGlue~\cite{sarlin2020superglue} requires two sets of keypoints and their visual descriptors as input and learns their correspondences using an attentional GNN with complete graphs over keypoints within and across images. Albeit the impressive performance, the computational and memory complexity of SuperGlue is quadratic to the number of keypoints, severely limiting its practicality to real-time or large-scale visual tasks. Some follow-ups try to address this problem by relaxing densely-connected graphs into sparsely-connected ones. Particularly, SGMNet~\cite{chen2021learning} which draws inspiration from guided matching methods~\cite{li2022guided} and graph pooling operations~\cite{ying2018hierarchical}, first generates a set of correspondences as seeds using an NN matcher, and then adopts a seeded GNN to perform message passing. Besides, ClusterGNN~\cite{shi2022clustergnn} adopts a learnable hierarchical clustering strategy to partition keypoints into different sub-graphs, and then performs message passing merely within these sub-graphs. In spite of improving the efficiency, both of them suffer from significant accuracy loss, as analyzed in Sec.~\ref{introduction}.


\subsection{Efficient Attention}
Transformer~\cite{vaswani2017attention} architectures have received extensive interest during the past few years. Particularly, in the context of attentional GNNs, Transformer serves as a graph-like model for broadcasting information across nodes in fully-connected graphs~\cite{wang2019learning}. In spite of its remarkable performance on a plethora of visual tasks~\cite{dosovitskiy2020image,carion2020end,wang2020axial,sun2021loftr}, the major drawback of Transformer is the quadratic complexity of the attention mechanism \emph{w.r.t.} the input size, significantly hampering its application to large-scale tasks.

In recent years, many works have spent great efforts to tackle the efficiency problem of attention mechanism. Specifically, they reduce computational complexity by the learning of partitioning or grouping on input elements~\cite{kitaev2020reformer,tay2020sparse}, linear projection functions~\cite{katharopoulos2020transformers,wang2020linformer}, or attention sharing~\cite{chen2021psvit}, to name just a few. Please refer to a comprehensive survey~\cite{tay2022efficient} for more details. Despite showing promising efficiency, such works typically concentrate on self-attention, where both queries and keys come from the same set of elements, and thus cannot be generalized well to the feature matching problem, which requires cross-attention to conduct cross-image communication.

\begin{figure*}[t]
	\centering
	\includegraphics[width=1\linewidth]{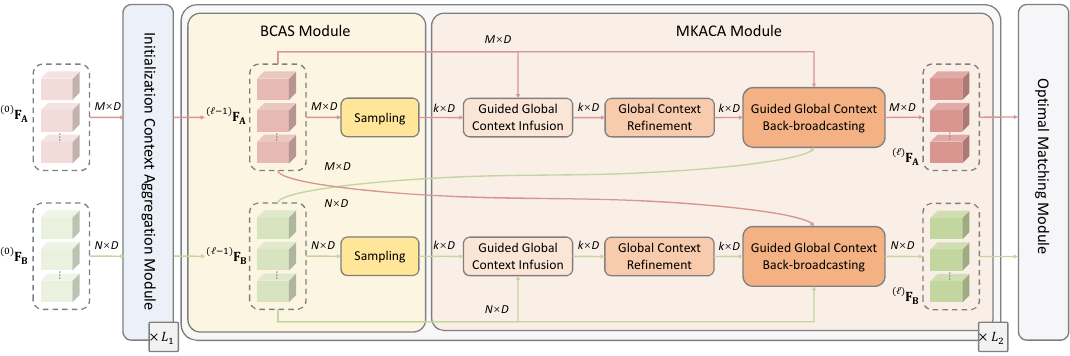}
    \vspace{-0.25in}
	\caption{The network architecture of MaKeGNN, which consists of four components. The initialization Context Aggregation (ICA) module (Sec.~\ref{ICA}) first performs coarse intra- and inter-graph message passing. Then the BCAS module (Sec.~\ref{BCAS}) and MKACA module (Sec.~\ref{MKACA}) reason about visual appearance similarity, neighborhood consensus, as well as the guidance provided by well-distributed matchable keypoint in the image pair jointly, to update per-node feature in a compact and robust manner. Finally, the optimal matching (OM) module (Sec.~\ref{OM}) utilizes Sinkhorn algorithm to derive the assignment matrix.}
	\label{framework}
\end{figure*}

\subsection{Outlier Rejection}
Among traditional methods, RANSAC and its variants~\cite{fischler1981random,raguram2012usac,barath2018graph,barath2020magsac++} are still the most popular outlier rejection methods. However, their major drawbacks are slow convergence and low accuracy in the case of high outlier rates. Due to the impressive performance achieved by the recent developments in deep learning, learning-based outlier rejection methods receive a continuous influx of attention. PointCN~\cite{yi2018learning} introduces Context Normalization to capture global contextual information. PointACN~\cite{sun2020acne} further improves the robustness of Context Normalization by weighing the importance of each feature map during the normalization step. OANet~\cite{zhang2019learning} maps matches to a set of clusters in a soft assignment manner, for local context exploring. CLNet~\cite{zhao2021progressive} adopts a local-to-global consensus learning
strategy to filter outliers progressively. LMCNet~\cite{liu2021learnable} and ConvMatch~\cite{zhang2023convmatch} learn the coherence and smoothness of the motion filed for outlier rejection. MS2DG-Net~\cite{dai2022ms2dg} injects semantic information into a GNN to extract local context. U-Match~\cite{li2023umatch} employs an encoder-decoder architecture, implicitly capturing the local context from different levels. However, these methods reason about visual appearance and geometric distribution separately and their performance strongly relies on the quality of candidate matches. Instead, our method can jointly process them and predict reliable correspondences end-to-end.

\section{Methodology}
We present Matchable Keypoint-Assisted Graph Neural Network, shorten as MaKeGNN, for learning reliable correspondences between two sets of keypoints with their corresponding visual descriptors. As illustrated in Fig.~\ref{framework}, the core modules of our framework are 1) \emph{Bilateral Context-Aware Sampling} (BCAS) \emph{Module} that dynamically samples two sets of well-distributed matchable keypoints from the image pair to guide the context aggregation process, and 2) \emph{Matchable Keypoint-Assisted Context Aggregation} (MKACA) \emph{Module} that considers sampled informative keypoints as attention bottlenecks for per-keypoint fine-grained feature update. In the following parts, we first present an overview of our network architecture, and then describe each module in detail.

%
\subsection{Overview}
We assume that a pair of images $\mathbf{A}$ and $\mathbf{B}$ depicting the same 3D structure have $M$ and $N$ keypoints and associated visual descriptors, respectively, which are indexed by $\alpha:=\{1,\dots,M\}$, $\beta:=\{1,\dots,N\}$, and extracted by off-the-shelf handcrafted features (\emph{e.g.}, SIFT~\cite{lowe2004distinctive} and ORB~\cite{rublee2011orb}) or learned ones (\emph{e.g.}, SuperPoint~\cite{detone2018superpoint} and R2D2~\cite{revaud2019r2d2}). The goal of our task is to establish geometrically consistent point-to-point correspondences between two images.

Following the \emph{de facto} standard~\cite{sarlin2020superglue,chen2021learning}, we formulate the feature matching task as a graph matching problem, where nodes are extracted keypoints from each image. Instead of employing an attentional GNN with complete graphs, we consider that matchable keypoints with the favorable property of being located in co-visible regions and having a true match in the other image can provide valuable matching priors. Thus we can dynamically sample two sets of well-distributed matchable keypoints from two images as message bottlenecks to guide the compact and robust context propagation across nodes in two graphs for subsequent matching. As we will show, this crucial difference increases not only accuracy but also efficiency.

An overview of MaKeGNN's workflow is illustrated in Fig.~\ref{framework}. Given two sets of local features $\{\mathbf{K}_\mathbf{A}^i\}_{i=1}^M$ and $\{\mathbf{K}_\mathbf{B}^i\}_{i=1}^N$ as input, $\mathbf{K}_\mathbf{I}^i=(\mathbf{p}_\mathbf{I}^i, \mathbf{d}_\mathbf{I}^i)$, where $\mathbf{I}\in\{\mathbf{A},\mathbf{B}\}$, $\mathbf{p}_\mathbf{I}^i=(x_\mathbf{I}^i,y_\mathbf{I}^i)\in\mathbb{R}^2$ is the image coordinate of keypoint $i$ in image $\mathbf{I}$, and $\mathbf{d}_\mathbf{I}^i\in\mathbb{R}^D$ a $D$ dimensional visual descriptor associated with keypoint $i$ in image $\mathbf{I}$, a position encoder first possesses them to obtain the initial representation vectors $^{(0)}\mathbf{F}^i_\mathbf{A}$ and $^{(0)}\mathbf{F}^i_\mathbf{B}$ by combining the visual appearance and location as~\cite{sarlin2020superglue,chen2021learning,shi2022clustergnn}:
\begin{equation}\label{eq1}
	^{(0)}\mathbf{F}^i_\mathbf{I} = \mathbf{d}_\mathbf{I}^i + f_{enc}(\mathbf{p}_\mathbf{I}^i), \;\mathbf{I}\in\{\mathbf{A},\mathbf{B}\},
\end{equation}
where $f_{enc}(\cdot)$ encodes the position of each keypoint into a high dimensional feature vector using an MLP. Then, $L_1$ \emph{Initialization Context Aggregation} (ICA) \emph{Modules} are stacked sequentially to perform coarse message passing, enabling each feature to roughly perceive intra- and inter-image contextual information. Subsequently, BCAS and MKACA modules that reason about visual appearance similarity, neighborhood consensus, as well as the guidance provided by well-distributed matchable keypoints in the image pair jointly, are stacked alternately $L_2$ times to achieve fine-grained feature update progressively. Finally, reliable correspondence are established by an \emph{Optimal Matching} (OM) \emph{Module} which formulates the assignment matrix using the Sinkhorn algorithm~\cite{cuturi2013sinkhorn}.

Next, we will provide the detailed description of our MakeGNN.

\subsection{Initialization Context Aggregation Module}\label{ICA}
Sampling two sets of well-distributed matchable keypoints from the image pair lays the foundation for subsequently compact and efficient message passing. However, it is impossible to directly predict the matchability score of each keypoint for sampling in light of the initial feature vectors $^{(0)}\mathbf{F}^i_\mathbf{A}$ and $^{(0)}\mathbf{F}^i_\mathbf{B}$, since they only encode intra-image geometric and visual cues and do not perceive inter-image ones. To this end, we introduce an ICA module, which enables the representation vector of each keypoint to roughly perceive intra- and inter-image contextual information. Specifically, similarly to~\cite{sarlin2020superglue,shi2022clustergnn}, the ICA module constructs complete graphs over keypoints within and across images, and lets them communicate with each other to retrieve coarse intra- and inter-image context for per-node feature update based on an attentional GNN~\cite{velivckovic2017graph}. This realizes the attentional aggregation using the standard attention mechanism~\cite{vaswani2017attention} combined with the feed-forward network (FFN) and shortcut connection. In particular, in a $D$-dimensional feature space, there are $M$ vectors, \emph{i.e.}, $\mathbf{X}\in\mathbb{R}^{M\times D}$, to be updated, $N$ vectors, \emph{i.e.}, $\mathbf{Y}\in\mathbb{R}^{N\times D}$, to be attended to, and an optional weight vector $\mathbf{w}\in\mathbb{R}^N$, to adjust the importance of each element in $\mathbf{Y}$. The attentional aggregation operation $\mathcal{G}(\cdot,\cdot,\cdot)$ is defined as:
\begin{equation}\label{eq2}
		\mathcal{G}(\mathbf{X},\mathbf{Y},\mathbf{w})=\mathbf{X}+\text{FFN}(\mathbf{X}||\mathcal{A}(\mathbf{X},\mathbf{Y},\mathbf{w})),
\end{equation}
where
\begin{equation}\label{eq3}
	\mathcal{A}(\mathbf{X},\mathbf{Y},\mathbf{w})=\theta(\mathbf{Q}\mathbf{K}^{\top})\mathbf{W}\mathbf{V},\mathbf{W}=\text{Diag}(\mathbf{w}),\mathbf{W}\in\mathbb{R}^{N\times N},
\end{equation}
$||$ means concatenation along row dimension, $\theta(\cdot)$ means Softmax operation, $\mathbf{Q}$ is linear projection of $\mathbf{X}$, and $\mathbf{K}$, $\mathbf{V}$ are linear projections of $\mathbf{Y}$.

Based on this definition, the initialization context aggregation process of ICA module in layer $\ell$ can be described as:
\begin{gather}\label{eq4}
		^{(\ell-1)}\mathbf{F}^1_\mathbf{I}=\mathcal{G}(^{(\ell-1)}\mathbf{F}_\mathbf{I},^{(\ell-1)}\mathbf{F}_\mathbf{I},\mathbf{1}),\nonumber\\
		^{(\ell)}\mathbf{F}_\mathbf{I}=\mathcal{G}(^{(\ell-1)}\mathbf{F}^1_\mathbf{I},^{(\ell-1)}\mathbf{F}^1_\mathbf{J},\mathbf{1}), \;\mathbf{I},\mathbf{J}\in\{\mathbf{A}, \mathbf{B}\}, \mathbf{I}\neq\mathbf{J},
\end{gather}
where $\mathbf{1}$ is an all-one vector, meaning that no weights are used in the aggregation process.

After being processed by several ICA modules, the updated feature vectors encode both the coarse visual and geometric context of each graph, and are fed into subsequent operations, \emph{i.e.}, BCAS and MKACA modules, for compact message passing and robust matching.

\subsection{Bilateral Context-Aware Sampling Module}\label{BCAS}
\begin{figure}[t]
	\centering
	\includegraphics[width=\linewidth]{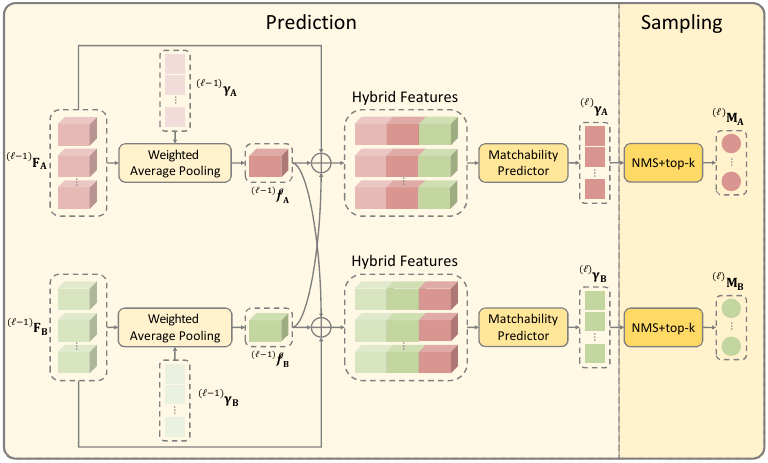}
    \vspace{-0.2in}
	\caption{Bilateral context-aware sampling module.}
	\label{sampler}
\end{figure}

As analyzed above, in the context of feature matching, there are a significant number of non-repeatable keypoints in the image pair due to occlusion and failure of the detector. Consequently, employing densely-connected graphs not only has no contributions to feature matching, but also introduces excessive redundant and pointless information interaction, which may impair the network's representation capability. To mitigate these problems, considering that matchable keypoints which are located in co-visible regions and have true matches in the other image, can provide valuable matching priors, we propose a sampling module that is capable of selecting well-distributed keypoints with high matchability scores dynamically, to guide the network for compact and meaningful message. Specifically, the proposed BCAS module consists of two procedures: matchability score prediction and matchable keypoint sampling, as presented in Fig.~\ref{sampler}.

\subsubsection{Matchability Score Prediction} 
\begin{figure}[t]
	\centering
	\includegraphics[width=0.8\linewidth]{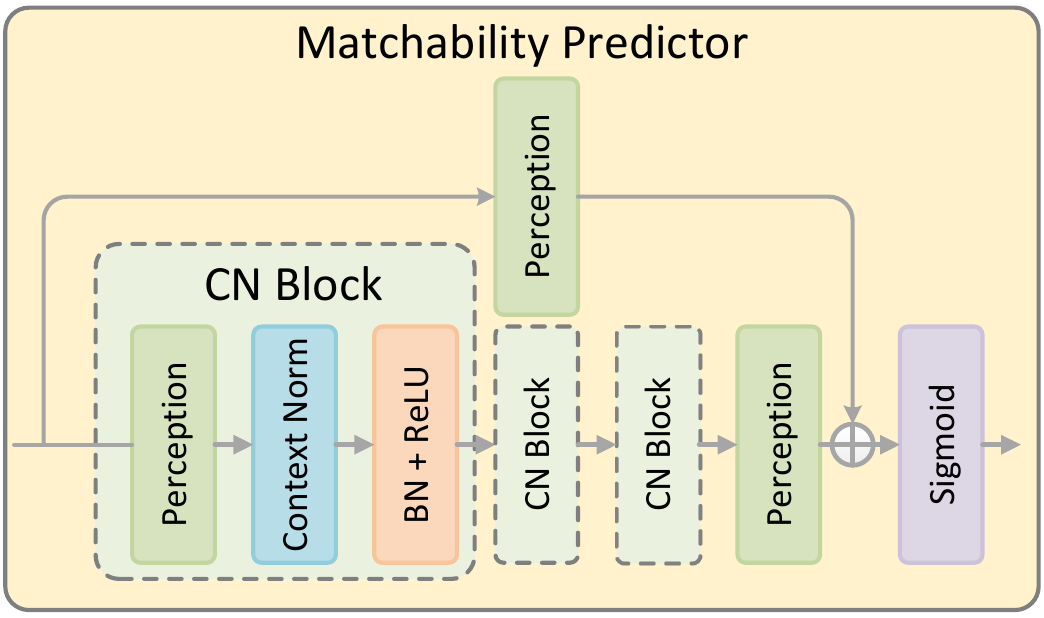}
	\caption{The structure of CN block and matchability predictor.}
	\label{CN}
\end{figure}

In the first stage, instead of relying solely on the feature of each keypoint for prediction, we also consider additional geometric information from both two images $\mathbf{A}$ and $\mathbf{B}$ to enhance the bilateral context awareness of per-keypoint feature, rendering the inferred matchability scores more reliable. In particular, given two sets of intermediate features $^{(\ell-1)}\mathbf{F}_\mathbf{A}$ and $^{(\ell-1)}\mathbf{F}_\mathbf{B}$, we could first generate the global representation vectors $^{(\ell-1)}\mathbf{f}_\mathbf{A}^g$, $^{(\ell-1)}\mathbf{f}_\mathbf{B}^g\in\mathbb{R}^D$ with a simple average pooling operation, which however is ill-suited for the feature matching task, since it overlooks the negative impact of a large portion of non-repeatable keypoints, making $^{(\ell-1)}\mathbf{f}_\mathbf{A}^g$ and $^{(\ell-1)}\mathbf{f}_\mathbf{B}^g$ not robust. To tackle this issue, we propose to weigh the importance of each keypoint based on the matchability scores $^{(\ell-1)}\boldsymbol{{\gamma}}_\mathbf{A}\in\mathbb{R}^M$ and $^{(\ell-1)}\boldsymbol{{\gamma}}_\mathbf{B}\in\mathbb{R}^N$ which are generated by the previous stage and elaborated later on, guiding the network to embed more robust global context. Following this train of thought, we extend average pooling
to a weighted formulation as follows:
\begin{equation}\label{eq5}
^{(\ell-1)}\mathbf{f}_\mathbf{I}^g=\textstyle{\sum_{i}}\theta(^{(\ell-1)}\boldsymbol{{\gamma}}_\mathbf{I})^i \cdot{^{(\ell-1)}\mathbf{F}_\mathbf{I}^i},\;\mathbf{I},\mathbf{J}\in\{\mathbf{A}, \mathbf{B}\},
\end{equation}
where $\theta$ means Softmax operation. Importantly, both $^{(L_1)}\boldsymbol{{\gamma}}_\mathbf{A}\in\mathbb{R}^M$ and $^{(L_1)}\boldsymbol{{\gamma}}_\mathbf{B}\in\mathbb{R}^N$ are all-one vectors. After obtaining $^{(\ell-1)}\mathbf{f}_\mathbf{A}^g$ and $^{(\ell-1)}\mathbf{f}_\mathbf{B}^g$, we concatenate them with per-keypoint feature accordingly, followed by a Context Normalization (CN)~\cite{yi2018learning} branch as the predictor to fuse the above hybrid features and predict new matchability scores $^{(\ell)}\boldsymbol{{\gamma}}_\mathbf{I}$ for each keypoint in image $\mathbf{I}\in\{\mathbf{A},\mathbf{B}\}$, which will be used to the select matchable keypoints in the later sampling stage. The prediction process can be summarized as follows:
\begin{equation}\label{eq6}
	^{(\ell)}\boldsymbol{{\gamma}}_\mathbf{I}^i=\phi(^{(\ell-1)}\mathbf{F}^i_\mathbf{I}||^{(\ell-1)}\mathbf{f}_\mathbf{I}^g||^{(\ell-1)}\mathbf{f}_\mathbf{J}^g),\;\mathbf{I},\mathbf{J}\in\{\mathbf{A}, \mathbf{B}\}, \mathbf{I}\neq\mathbf{J},
\end{equation}
where $\phi(\cdot)$ (\emph{i.e.}, matchability predictor) is comprised of lightweight stacked CN~\cite{yi2018learning} blocks, whose detailed structure is reported in Fig.~\ref{CN}. The input/output channel numbers of five MLPs are 3$D$/3$D$, 3$D$/$D$, $D$/$D$, $D$/1, 3$D$/1, respectively, where $D$ denotes the dimension of initial visual descriptors.

\subsubsection{Matchable Keypoint Sampling} In the second stage, keypoints in each image with top-\textit{k} matchability scores are sampled as candidates, to provide guidance for message passing. Also, we employ a simple yet effective Non-Maximum Suppression (NMS) post-process to ensure a better spatial distribution of selected keypoints. The NMS's radius for image $\mathbf{A}$ is $\frac{\sigma}{\vert\alpha\vert(\vert\alpha\vert-1)}\sum\limits_{(i,j)\in\alpha\times\alpha,i\neq j}d^{ij}_\mathbf{A}$, where $\alpha$ is the index list for all keypoints in image $\mathbf{A}$, $d^{ij}_\mathbf{A}$ is the distance between keypoint $i$ and $j$ in image $\mathbf{A}$, $\vert\cdot\vert$ denotes the cardinality of a set, and $\sigma$ controls the radius, which is set to $5\times10^{-2}$ for all experiments.

The BCAS module at the $\ell$-th layer outputs $^{(\ell)}\boldsymbol{{\gamma}}=(^{(\ell)}\boldsymbol{{\gamma}}_\mathbf{A},^{(\ell)}\boldsymbol{{\gamma}}_\mathbf{B})$, where $^{(\ell)}\boldsymbol{{\gamma}}_\mathbf{A}\in[0,1]^M$ and $^{(\ell)}\boldsymbol{{\gamma}}_\mathbf{B}\in[0,1]^N$ are matchability score sets for keypoints in each image, and $^{(\ell)}\mathbf{M}=(^{(\ell)}\mathbf{M}_\mathbf{A},^{(\ell)}\mathbf{M}_\mathbf{B})$, where $^{(\ell)}\mathbf{M}_\mathbf{A}$ and $^{(\ell)}\mathbf{M}_\mathbf{B}$ are index lists for well-distributed matchable keypoints in each image.

\subsection{Matchable Keypoint-Assisted Context Aggregation Module}\label{MKACA}
As discussed above, the MKACA module utilizes selected well-distributed matchable keypoints as attention bottlenecks, so that each keypoint only exchanges messages with intra- and inter- bottlenecks for compact and robust context aggregation. To this end, we adopt an infusion-enhancement-broadcasting strategy in each MKACA module, fleshed out by the following three operations.

\subsubsection{Guided Global Context Infusion} Assuming that the MKACA module in layer $\ell$~\footnote{It should be mentioned that the proposed BCAS and MKACA modules are inseparable. Therefore, we group a BCAS module followed by an MKACA module into the same GNN layer and regard them as a processing unit.} takes features $^{(\ell-1)}\mathbf{F}_\mathbf{A}$ and $^{(\ell-1)}\mathbf{F}_\mathbf{B}$ with index sets $^{(\ell)}\mathbf{M}=(^{(\ell)}\mathbf{M}_\mathbf{A},^{(\ell)}\mathbf{M}_\mathbf{B})$ as input, the features for matchable keypoints, \emph{i.e.}, $_\mathbf{M}^{(\ell)}\mathbf{F}_\mathbf{A}$, $_\mathbf{M}^{(\ell)}\mathbf{F}_\mathbf{B}\in\mathbb{R}^{\textit{k}\times D}$, are first retrieved by $^{(\ell)}\mathbf{M}$:
\begin{equation}\label{eq7}
_\mathbf{M}^{(\ell)}\mathbf{F}_\mathbf{I}=^{(\ell-1)}\mathbf{F}_\mathbf{I}[^{(\ell)}\mathbf{M}_\mathbf{I}],\;\mathbf{I},\mathbf{J}\in\{\mathbf{A}, \mathbf{B}\},
\end{equation}
where $[\cdot]$ denotes the indexing operation. Then, we infuse contextual information from every keypoint in each graph into intra-matchable ones through a matchabilty score-guided attentional aggreagtion, which enables valuable matching priors to guide the gathering process, suppressing interference of considerable non-repeatable keypoints caused by occlusion and failure of the detector, contributing to the robustness of our network to the potential noise in initial keypoints, and resulting in cleaner and sharper per-node feature update:
\begin{equation}\label{eq8}
	_\mathbf{M}^{(\ell)}\mathbf{F}_\mathbf{I}^1=\mathcal{G}(_\mathbf{M}^{(\ell)}\mathbf{F}_\mathbf{I},^{(\ell-1)}\mathbf{F}_\mathbf{I},^{(\ell)}\boldsymbol{{\gamma}}_\mathbf{I}),\;\mathbf{I},\mathbf{J}\in\{\mathbf{A}, \mathbf{B}\}.
\end{equation}
By doing so, the updated features $_\mathbf{M}^{(\ell)}\mathbf{F}_\mathbf{A}^1$ and $_\mathbf{M}^{(\ell)}\mathbf{F}_\mathbf{B}^1$ encode both visual and geometric context from their respective corresponding graphs.

\subsubsection{Global Context Refinement} To further enhance the encoded global context, we propose to conduct intra-graph information interaction among well-distributed matchable keypoints. More specifically, the intra-graph attentional aggregation is applied to the features of matchable keypoints $_\mathbf{M}^{(\ell)}\mathbf{F}_\mathbf{I}^1$ for global context refinement:
\begin{equation}\label{eq9}
	_\mathbf{M}^{(\ell)}\mathbf{F}_\mathbf{I}^2=\mathcal{G}(_\mathbf{M}^{(\ell)}\mathbf{F}_\mathbf{I}^1,_\mathbf{M}^{(\ell)}\mathbf{F}_\mathbf{I}^1,\mathbf{1}),\;\mathbf{I},\mathbf{J}\in\{\mathbf{A}, \mathbf{B}\},
\end{equation}
where $_\mathbf{M}^{(\ell)}\mathbf{F}_\mathbf{A}^2$ and $_\mathbf{M}^{(\ell)}\mathbf{F}_\mathbf{B}^2$ are the features of matchable keypoints with enhanced global context.

\subsubsection{Guided Global Context Back-broadcasting} After the intra-graph message interaction among matchable keypoints, every keypoint on each side retrieves fine-grained global context from both intra-matchable and inter-matchable keypoints through a matchabilty score-guided attentional aggreagtion:
\begin{gather}\label{eq10}
	^{(\ell-1)}\mathbf{F}^1_\mathbf{I}=\mathcal{G}(^{(\ell-1)}\mathbf{F}_\mathbf{I},_\mathbf{M}^{(\ell)}\mathbf{F}_\mathbf{I}^2,_\mathbf{M}^{(\ell)}\boldsymbol{{\gamma}}_\mathbf{I}),\nonumber\\
	^{(\ell)}\mathbf{F}_\mathbf{I}\!=\!\mathcal{G}(^{(\ell-1)}\mathbf{F}^1_\mathbf{I},_\mathbf{M}^{(\ell)}\mathbf{F}_\mathbf{J}^2,_\mathbf{M}^{(\ell)}\boldsymbol{{\gamma}}_\mathbf{J}), \;\mathbf{I},\mathbf{J}\in\{\mathbf{A}, \mathbf{B}\}, \mathbf{I}\!\neq\!\mathbf{J},
\end{gather}
where $_\mathbf{M}^{(\ell)}\boldsymbol{{\gamma}}_\mathbf{I}=^{(\ell)}\boldsymbol{{\gamma}}_\mathbf{I}[^{(\ell)}\mathbf{M}_\mathbf{I}]$.

It should be mentioned that leveraging the matchabilty scores $_\mathbf{M}^{(\ell)}\boldsymbol{{\gamma}}_\mathbf{A}$ and $_\mathbf{M}^{(\ell)}\boldsymbol{{\gamma}}_\mathbf{B}$ into the backpropagation process can suppress adverse message passing from mis-sampled matchable keypoints, ensuring the robustness of our network to the potential noise in bottlenecks.

The MKACA module at the $\ell$-th layer outputs the updated keypoint features $^{(\ell)}\mathbf{F}_\mathbf{A}$ and $^{(\ell)}\mathbf{F}_\mathbf{B}$ with favorable visual and geometric context from both images $\mathbf{A}$ and $\mathbf{B}$, conducive to subsequent matching.

\subsection{Optimal Matching Module}\label{OM}
After passing through $L$ processing units, we can obtain the final keypoint features of the image pair, \emph{i.e.}, $^{(L)}\mathbf{F}_\mathbf{A}\in\mathbb{R}^{M\times D}$ and $^{(L)}\mathbf{F}_\mathbf{B}\in\mathbb{R}^{N\times D}$. Following SuperGlue~\cite{sarlin2020superglue}, we first compute the similarity matrix ${\mathbf{S}=} ^{(L)}\mathbf{F}_\mathbf{A} ^{(L)}\mathbf{F}_\mathbf{B}^\top$ and then augment $\mathbf{S}$ to $\hat{\mathbf{S}}$ by appending a new row and column as dustbins, filled with a single learnable parameter $z$. Finally, Sinkhorn~\cite{cuturi2013sinkhorn} algorithm is applied on $\hat{\mathbf{S}}$ to generate the assignment matrix $\mathbf{P}\in[0,1]^{M\times N}$:
\begin{gather}\label{eq11_12}
\hat{S}_{i,j}=\left\{
\begin{aligned}
	&{S}_{i,j}, \;i\leq M, j\leq N \\
	&z, \;\text{otherwise} \\
\end{aligned}, \; \hat{\mathbf{S}}\in\mathbb{R}^{(M+1)\times(N+1)},
\right. \\
\hat{\mathbf{P}}=\text{Sinkhorn}(\hat{\mathbf{S}}),\; \mathbf{P}=\hat{\mathbf{P}}_{1:M,1:N},
\end{gather}
where $\mathbf{P}=\hat{\mathbf{P}}_{1:M,1:N}$ denotes the first $M$ rows and $N$ columns of matrix $\hat{\mathbf{P}}$.

After deriving the assignment matrix $\mathbf{P}$, the reliable matches can be established by an MNN check with a predefined inlier threshold.

\subsection{Loss Formulation}
We formulate our loss as two parts: 1) \emph{Matching Loss}, which enforces the network to predict as many correct matches as possible, and 2) \emph{Binary Classification Loss}, which guides the BCAS module to produce reliable matchability scores, ensuring that the sampled keypoints are matchable. Given the indices of ground truth matches $\mathbf{C}_\mathbf{m}=\{(i,j)\}\in\alpha\times\beta$ and non-repeatable keypoints $\mathbf{C}_{\mathbf{nA}}$, $\mathbf{C}_{\mathbf{nB}}$, where a keypoint is regarded as non-repeatable (unmatchable) if there are no corresponding keypoints in the other image, our \emph{Matching Loss} $\mathcal{L}_{match}$ is defined as:
\begin{align}\label{eq13}
&\mathcal{L}_{match}=-\frac{1}{\vert\mathbf{C}_\mathbf{m}\vert}\sum_{(i,j)\in\mathbf{C}_\mathbf{m}}\log(\hat{\mathbf{P}}_{i,j})\nonumber\\
&-\frac{1}{2\vert\mathbf{C}_\mathbf{nA}\vert}\sum_{i\in\mathbf{C}_{\mathbf{nA}}}\log(\hat{\mathbf{P}}_{i,N+1})-\frac{1}{2\vert\mathbf{C}_\mathbf{nB}\vert}\sum_{j\in\mathbf{C}_{\mathbf{nB}}}\log(\hat{\mathbf{P}}_{M+1,j}).
\end{align}

As for the \emph{Binary Classification Loss} $\mathcal{L}_{cls}$ that is used to train the matchability predictor in Sec.~\ref{BCAS}, we adopt a cross entropy loss for matchable/non-repeatable keypoint binary classification. Specifically, a keypoint whose index belongs to the index list $\mathbf{C}_{\mathbf{nA}}$ or $\mathbf{C}_{\mathbf{nB}}$ will be labeled as 0 (\emph{i.e.}, non-repeatable), otherwise it will be labeled as 1 (\emph{i.e.}, matchable keypoints). Details of $\mathcal{L}_{cls}$ can refer to some learning-based outlier rejection methods~\cite{zhang2019learning}.

To sum up, the optimization objective of our network is to minimize a hybrid loss function as follows:
\begin{equation}\label{eq14}
\mathcal{L}=\mathcal{L}_{match}+\lambda\sum_{\ell=L_1+1}^{L}{^{(\ell)}\mathcal{L}_{cls}},
\end{equation}
where ${^{(\ell)}\mathcal{L}_{cls}}$ means the \emph{Binary Classification Loss} of $\ell$-th processing unit, $L_1$ and $L$ are the total number of initialization processing units and GNN layers cascaded in the network, respectively, and $\lambda$ is a weight to balance the two loss terms.

\section{Experiments}
In the following sessions, we first present the implementation details of our MaKeGNN, then evaluate it on three diverse tasks which heavily rely on the quality of matches established by feature matching methods, namely: relative camera estimation, fundamental matrix estimation, and visual localization. Finally, a comprehensive analysis is provided for better understanding MaKeGNN.
\subsection{Implementation Details}
Following SGMNet~\cite{chen2021learning}, we train our network on GL3D dataset~\cite{shen2019matchable} from
scratch, which covers outdoor and indoor scenes, to obtain a generic model. Concretely, GL3D is originally based on 3D reconstruction of 543 different scenes including landmarks and small objects, while in its latest version additional 713 sequences of internet tourism photos are added. For each sequence, we sample 1$k$ pairs of images and retain those with common track ration in the range of $[0.1, 0.5]$ and rotation angle in the range of $[6^\circ,60^\circ]$, both of which are available in the original dataset, making sure that the image pairs for training are not too hard or too easy. Also, we reproject each keypoint to the other image and compute its reprojection distances with depth maps, to obtain the indices of ground truth matches and non-repeatable keypoints. To be specific, a keypoint is labeled as non-repeatable if its reprojection distances with all keypoints in the other image are larger than 10 pixels, while a pair of keypoints that are nearest to each other after reprojection and with a reprojection distance lower than 3 pixels, are considered as ground truth matches, and both of the two keypoints are regraded to be matchable. We further remove image pairs which contains fewer than 50 ground truth matches. Finally, about 400$k$ image pairs are generated for training.

We sample 1$k$ keypoints and 128 well-distributed keypoints with high matchability scores for each image during the training process. We adopt Adam optimizer with a learning rate of $10^{-4}$ and a batchsize of 16 in optimization. We utilize a learning decay strategy after 300$k$ iterations with a decay rate of 0.999996 until 900$k$ iterations. Matchability score weight $\lambda$ in loss is set to 5. The total number of processing units is set to $L=9$ ($L_1=3$ and $L_2=6$). We use 4-head attention in both ICA and MKACA modules, and perform 100 Sinkhorn iterations to obtain the assignment matrix. During the testing phase, we sample $[\frac{128\#\text{keypoint}}{2000}]$ matchable keypoints, where $\#\text{keypoint}$ denotes the number of keypoints, and use an inlier threshold of 0.2 to keep reliable matches from the soft assignment. All experiments are conducted on Ubuntu 18.04 with
GeForce RTX 3090 GPUs.

\subsection{Relative Pose Estimation}
Relative pose estimation refers to estimate the relative position relationship (\emph{i.e.}, rotation and translation) between different camera views, the accuracy of which heavily relies on the quality of model-predicted matching pairs.

\subsubsection{Datasets} As in the previous work~\cite{chen2021learning}, we adopt two popular datasets, YFCC100M~\cite{thomee2016yfcc100m} and ScanNet~\cite{dai2017bundlefusion} to evaluate the feature matching ability of our MaKeGNN on this task in outdoor and indoor scenes, respectively. More specifically, for YFCC100M which is a large-scale outdoor dataset containing 100 million images from Internet, we leverage camera poses and sparse 3D models provided by COLMAP~\cite{schonberger2016structure} to generate ground truth, and choose 4 sequences for evaluation as in~\cite{zhang2019learning}. ScanNet which is a large-scale indoor dataset widely used for depth prediction~\cite{bae2022multi}, consists of monocular sequences with ground truth poses and depth images. Following SuperGlue~\cite{sarlin2020superglue}, we select 1500 image pairs for evaluation. We evaluate MaKeGNN using both hand-designed RootSIFT~\cite{arandjelovic2012three,lowe2004distinctive} and learning-based SuperPoint~\cite{detone2018superpoint} features. Particularly, we detect up to 2$k$ keypoints on YFCC100M for both features, and up to 1$k$ and 2$k$ keypoints on ScanNet for SuperPoint and RootSIFT, respectively.

\subsubsection{Evaluation Protocols} Following~\cite{sarlin2020superglue,chen2021learning,shi2022clustergnn}, we report the area under the cumulative error curve (AUC) of the pose error (\emph{i.e.}, the maximum of the angular error in rotation and translation) at multiple thresholds ($5^{\circ}$, $10^{\circ}$, $20^{\circ}$). Importantly, relative poses are recovered by estimating essential matrix with RANSAC. We also report the mean matching score (M.S.)~\cite{detone2018superpoint,sarlin2020superglue}, \emph{i.e.}, the ratio of correct matches to total keypoints, and the mean precision (Prec.)~\cite{detone2018superpoint,sarlin2020superglue} of the established matches, to evaluate the matching performance.

\subsubsection{Baselines} We compare MaKeGNN with 1) traditional heuristic pruning strategies, including NN+RT, MNN~\cite{lowe2004distinctive}, and AdaLAM~\cite{cavalli2020handcrafted}, and 2) learning-based matching methods, including OANet~\cite{zhang2019learning} and ConvMatch~\cite{zhang2023convmatch}, SuperGlue~\cite{sarlin2020superglue}, SGMNet~\cite{chen2021learning}, and ParaFormer~\cite{lu2023paraformer}. For the sake of fairness, we meticulously retrain all the learning-based methods on the same sequences of GL3D~\cite{shen2019matchable} where 1$k$ keypoints are detected per image.

\subsubsection{Results of Outdoor Pose Estimation} 
\begin{figure}[t]
	\centering
	\includegraphics[width=\linewidth]{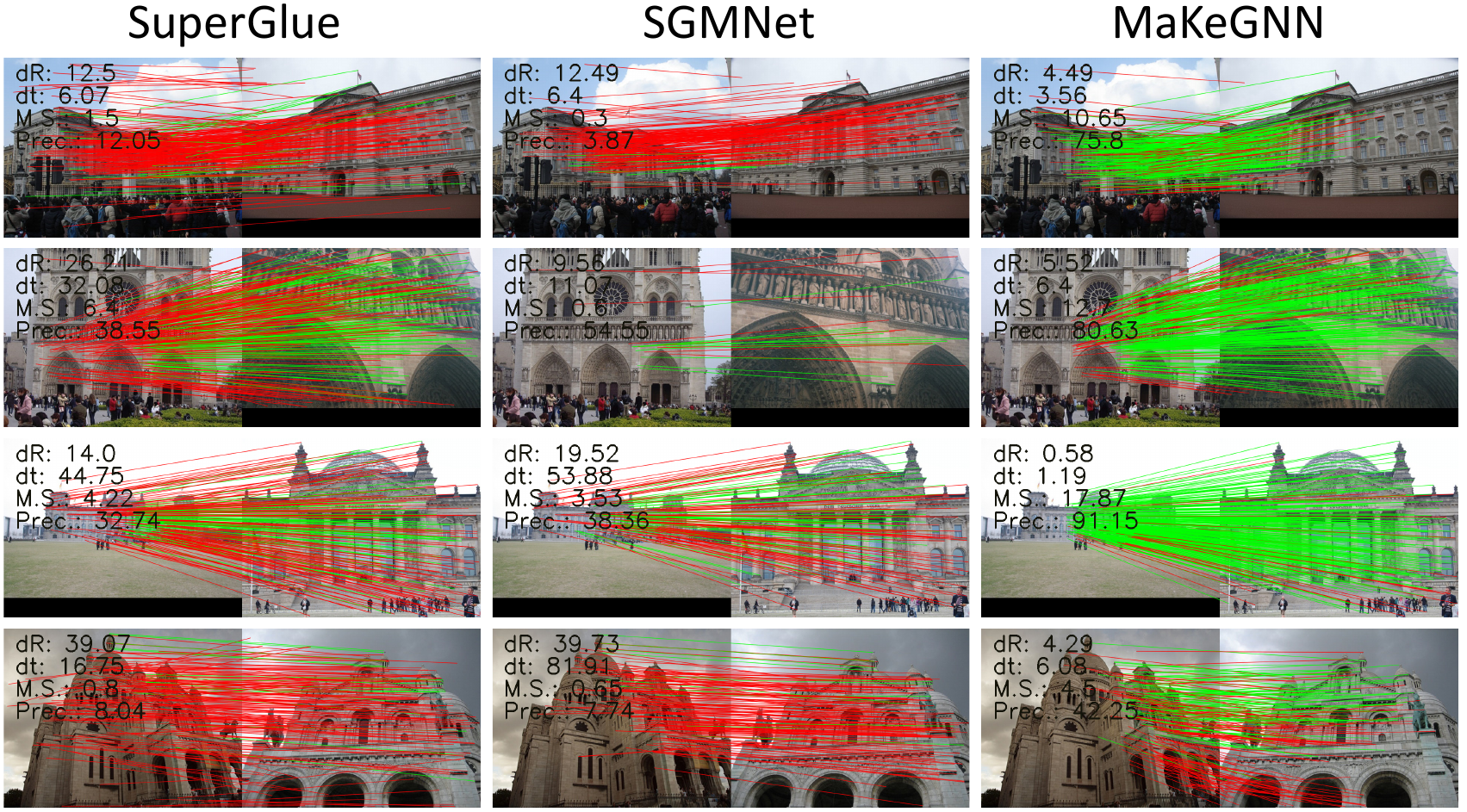}
	\caption{Matching results on YFCC100M with SuperPoint features. Correspondences are in \textcolor{green}{green} if they are consistent with the
		ground truth epipolar geometry, and in \textcolor{red}{red} otherwise. Pose estimation results, M.S., and Prec. are shown at the top left corner.}
	\label{outdoor_pose}
\end{figure}

The challenges of outdoor image matching mainly come from large illumination variation, extreme viewpoint change, and occlusion. We present several visualization results of outdoor scenes in Fig.~\ref{outdoor_pose}. As is evident, in comparison to SuperGlue and SGMNet, our MaKeGNN can retrieve richer geometrically consistent matches with fewer mismatches.

The quantitative results with RANSAC post-processing are reported in Tab.~\ref{yfcc}. It is worth noting that when using SuperPoint for evaluation, we adopt MNN as the comparative method instead of NN+RT, as the latter would establish considerably fewer matches~\cite{detone2018superpoint}. Undoubtedly, the performance of such heuristic tricks, \emph{i.e.}, NN+RT and MNN, is quite poor. While outlier rejection methods, AdaLAM and OANet, can greatly improve the precision of initial matches, their performance is also limited by the quality of the initial set. Clearly, compared to these above methods, learning-based matchers achieve significant higher pose accuracy. In particular, our MaKeGNN consistently outperforms other competitors on pose estimation at all threshold and delivers competitive matching performance regardless of using RootSIFT or SuperPoint, due to the compact and robust message passing guided by well-distributed matchable keypoints.

\begin{table}[t]
	\caption{Evaluation on YFCC100M for outdoor pose estimation. Pose estimation AUC at different thresholds, M.S., and Prec. are reported. \textcolor[RGB]{254,67,101}{Red} indicates the best, \textcolor[RGB]{0,124,62}{green} ranks the second, and \textcolor[RGB]{0,90,171}{blue} ranks the third.}
	\label{yfcc}
	\centering
	\resizebox{\linewidth}{!}{
	\begin{tabular}{ccccccc}
		\toprule
		\multirow{2}{*}{Feature}   & \multirow{2}{*}{Matcher}       & \multicolumn{3}{c}{Pose estimation AUC} & \multirow{2}{*}{M.S.} & \multirow{2}{*}{Prec.}
		\\
		\cmidrule{3-5}&&@$5^{\circ}$&@$10^{\circ}$&@$20^{\circ}$&&\\
		\midrule
		\multirow{8}{*}{RootSIFT}
		& NN+RT& 48.53& 58.16  &68.13 & 4.44& 56.38 \\
		& AdaLAM (4$k$) & 57.78& 68.01&77.38& 7.92& \textcolor[RGB]{0,90,171}{83.15}                           \\
		& OANet& 58.00& 67.80 &77.46& 5.84& 81.80                            \\
		& ConvMatch& 59.23& 68.54 &77.56& 8.03& 67.91                            \\
		& SuperGlue&  \textcolor[RGB]{0,124,62}{63.20}&  \textcolor[RGB]{0,124,62}{72.71} & \textcolor[RGB]{0,124,62}{81.86}& 16.60& 81.71                           \\
		& SGMNet& \textcolor[RGB]{0,90,171}{62.80}  & \textcolor[RGB]{0,90,171}{72.49}   &\textcolor[RGB]{0,90,171}{81.45}  & \textcolor[RGB]{0,90,171}{17.06}  & \textcolor[RGB]{254,67,101}{86.09}                           \\
		& ParaFormer&61.80&71.70&81.18& \textcolor[RGB]{0,124,62}{17.09}&76.29\\
		& MaKeGNN&\textcolor[RGB]{254,67,101}{64.73}  &\textcolor[RGB]{254,67,101}{73.91}  &\textcolor[RGB]{254,67,101}{82.78} &\textcolor[RGB]{254,67,101}{17.54} & \textcolor[RGB]{0,124,62}{84.59} \\
		\midrule
		\multirow{8}{*}{SuperPoint}
		&MNN&35.83&44.18&53.66&3.63& 58.96\\
		&AdaLAM (2$k$)&40.20&49.03&59.11&10.17& 72.57\\
		&OANet&48.80&59.06&70.02&12.48&71.95\\
		&ConvMatch& 50.28& 61.25 &71.78&12.58 & 68.34\\
		&SuperGlue&60.25&70.46&80.06& 19.29&78.19\\
		&SGMNet&\textcolor[RGB]{0,90,171}{61.13}&\textcolor[RGB]{0,90,171}{71.06}&\textcolor[RGB]{0,90,171}{80.18}& \textcolor[RGB]{0,90,171}{22.06}&\textcolor[RGB]{0,124,62}{84.61}\\
		&ParaFormer&\textcolor[RGB]{0,124,62}{61.75}&\textcolor[RGB]{0,124,62}{72.03}&\textcolor[RGB]{0,124,62}{81.23}&\textcolor[RGB]{0,124,62}{22.31}&\textcolor[RGB]{0,90,171}{81.28}\\
		&MaKeGNN&\textcolor[RGB]{254,67,101}{63.70}&\textcolor[RGB]{254,67,101}{73.48}&\textcolor[RGB]{254,67,101}{82.22}&\textcolor[RGB]{254,67,101}{23.04}&\textcolor[RGB]{254,67,101}{85.58}\\
		\bottomrule
	\end{tabular}}
\end{table}

\subsubsection{Results of Indoor Pose Estimation} 
\begin{figure}[t]
	\centering
	\includegraphics[width=\linewidth]{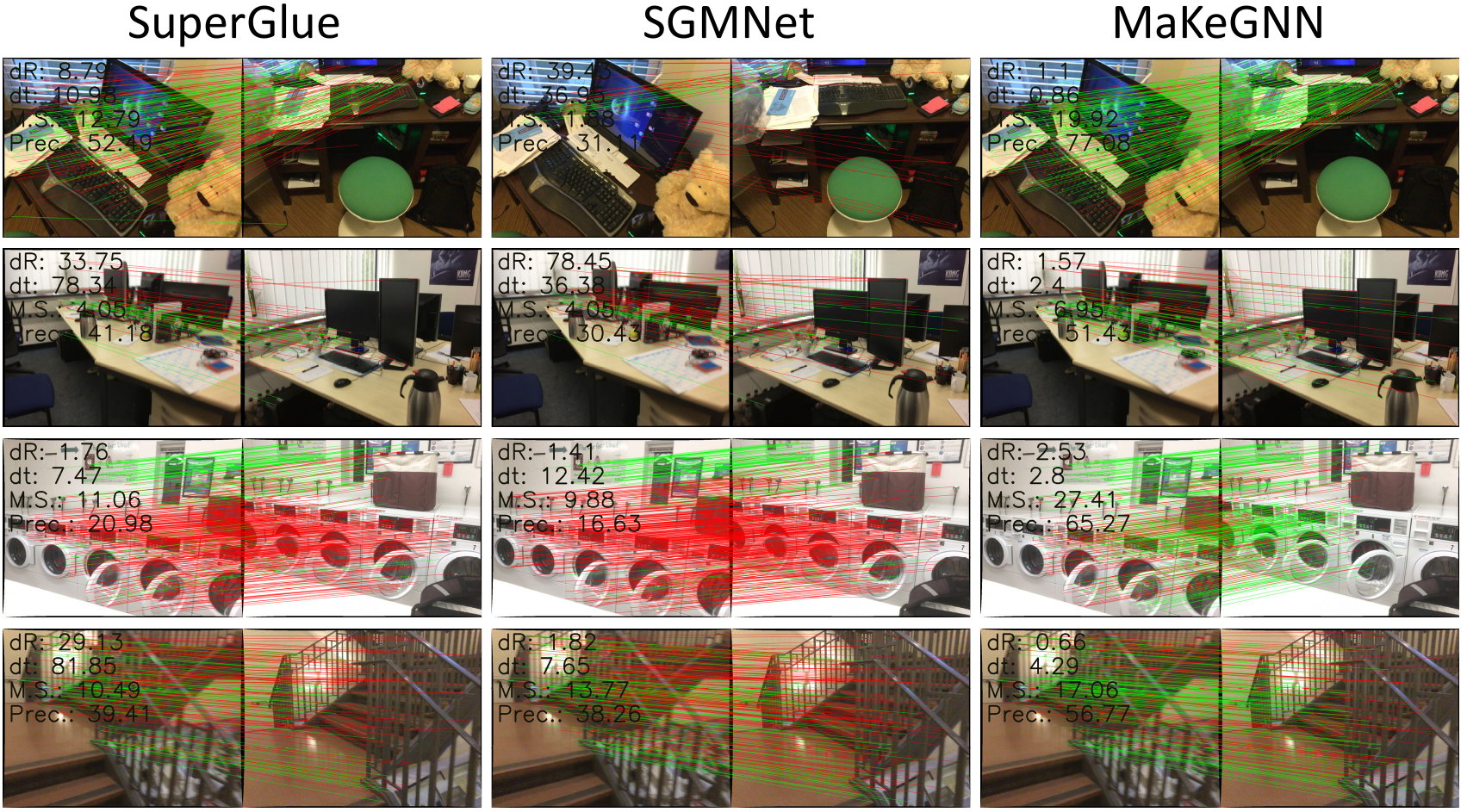}
	\caption{Matching results on ScanNet with SuperPoint features. Correspondences are in \textcolor{green}{green} if they are consistent with the
		ground truth epipolar geometry, and in \textcolor{red}{red} otherwise. Pose estimation results, M.S., and Prec. are shown at the top left corner.}
	\label{indoor_pose}
\end{figure}

We also evaluate the performance of our MaKeGNN on ScanNet for indoor pose estimation, which presents greater challenges compared to outdoor scenes due to factors such as poor texture and repetitive elements.
Even so, visualization results shown in Fig.~\ref{indoor_pose} qualitatively demonstrate that MakeGNN can achieve impressive performance. Further looking at Tab.~\ref{ScanNet}, as with YFCC100M, MakeGNN is still more powerful than other baselines at all AUC thresholds with both RootSIFT and SuperPoint, indicating that well-distributed matchable keypoints can provide valuable matching priors to guide the context aggregation process, suppressing interference of considerable non-repeatable keypoints and enabling accurate pose estimation.

\begin{table}[t]
	\caption{Evaluation on ScanNet for indoor pose estimation. Pose estimation AUC at different thresholds, M.S., and Prec. are reported. \textcolor[RGB]{254,67,101}{Red} indicates the best, \textcolor[RGB]{0,124,62}{green} ranks the second, and \textcolor[RGB]{0,90,171}{blue} ranks the third.}
	\label{ScanNet}
	\centering
	\resizebox{\linewidth}{!}{
		\begin{tabular}{ccccccc}
			\toprule
			\multirow{2}{*}{Feature}   & \multirow{2}{*}{Matcher}       & \multicolumn{3}{c}{Pose estimation AUC} & \multirow{2}{*}{M.S.} & \multirow{2}{*}{Prec.}
			\\
			\cmidrule{3-5}&&@$5^{\circ}$&@$10^{\circ}$&@$20^{\circ}$&&\\
			\midrule
			\multirow{6}{*}{RootSIFT}
			& NN+RT& 21.59& 30.82  &40.56 & 2.33& 28.26 \\
			& ConvMatch& 24.76& 33.43 &43.02& 4.28& 32.20                            \\
			& SuperGlue& \textcolor[RGB]{0,90,171}{32.47}& 42.53 &53.20& 8.43& \textcolor[RGB]{0,90,171}{42.22}   \\       
			& SGMNet& 32.29  & \textcolor[RGB]{0,90,171}{42.81}   &\textcolor[RGB]{0,90,171}{54.13}  & \textcolor[RGB]{0,90,171}{8.80}  & \textcolor[RGB]{254,67,101}{45.59}                        \\
			& ParaFormer&\textcolor[RGB]{0,124,62}{32.62}&\textcolor[RGB]{0,124,62}{43.32}&\textcolor[RGB]{0,124,62}{54.21}&\textcolor[RGB]{0,124,62}{9.09}&38.38\\
			& MaKeGNN&\textcolor[RGB]{254,67,101}{33.69} &\textcolor[RGB]{254,67,101}{44.05}&\textcolor[RGB]{254,67,101}{55.13} &\textcolor[RGB]{254,67,101}{9.29} &\textcolor[RGB]{0,124,62}{44.68}  \\
			\midrule
			\multirow{6}{*}{SuperPoint}
			&MNN&19.72&27.81&38.00&8.72&44.54 \\
			&ConvMatch& 24.63& 33.93 &40.25& 14.94& 37.37                            \\
			&SuperGlue&\textcolor[RGB]{0,90,171}{34.18}&\textcolor[RGB]{0,90,171}{44.35}&\textcolor[RGB]{0,90,171}{54.89}& 16.25&46.16\\
			&SGMNet&32.42&42.88&54.58&\textcolor[RGB]{0,124,62}{16.63}&\textcolor[RGB]{0,124,62}{47.63}\\
			&ParaFormer&\textcolor[RGB]{0,124,62}{34.96}&\textcolor[RGB]{0,124,62}{45.66}&\textcolor[RGB]{0,124,62}{56.85}&\textcolor[RGB]{0,90,171}{16.62}&\textcolor[RGB]{0,90,171}{46.64}\\
			&MaKeGNN&\textcolor[RGB]{254,67,101}{36.43}&\textcolor[RGB]{254,67,101}{47.06}&\textcolor[RGB]{254,67,101}{58.04}&\textcolor[RGB]{254,67,101}{16.88}&\textcolor[RGB]{254,67,101}{48.39}\\
			\bottomrule
	\end{tabular}}
\end{table}

\subsection{Fundamental Matrix Estimation}
We further apply our method to recover the fundamental matrix (FM) with uncalibrated cameras, which suits more general-purpose scenes than other geometric models, \emph{e.g.}, homography and essential matrix.

\subsubsection{Dataset} We resort to FM-Bench~\cite{bian2019evaluation} to validate the effectiveness of our network on fundamental matrix estimation. More specifically, FM-Bench is comprised of four sub-datasets: TUM~\cite{sturm2012benchmark}, KITTI~\cite{geiger2012we}, Tanks and Temples (T\&T)~\cite{knapitsch2017tanks}, and Community Photo Collection (CPC)~\cite{wilson2014robust}, the details of which are summarized in Tab.~\ref{FM_datasets}. We randomly select
1000 matched image pairs from each dataset for evaluation. In particular, TUM and KITTI provide ground truth camera intrinsics and extrinsics, enabling us to directly derive the ground truth fundamental matrix from them. However, for T\&T and CPC, where camera intrinsics and extrinsics are unknown, we apply COLMAP~\cite{schonberger2016structure} to reconstruct the image sequences and consequently obtain the ground truth matrix.
\begin{table*}[t]
	\centering
	\caption{Details of the experimental datasets for fundamental matrix estimation.}\label{FM_datasets}

	\begin{tabular}{cccccc}
		\toprule
		Dataset&Description&\#Seq&\#Image&Resolution&Baseline\\
		\midrule
		TUM~\cite{sturm2012benchmark}&indoor scenes&3&5994&480$\times$640&short\\
		
		KITTI~\cite{geiger2012we} &street views&5&9065&370$\times$1226&short\\
		
		T\&T~\cite{knapitsch2017tanks} &outdoor scenes&3&922&1080$\times$2048  1080$\times$1920&wide\\
		
		CPC~\cite{wilson2014robust} &internet photos&1&1615&varying&wide\\
		\bottomrule
	\end{tabular}
\end{table*}

\subsubsection{Evaluation Protocols} After establishing matches with feature matching approaches, we can estimate the fundamental matrix for each image pair with RANSAC. Considering the different image resolutions, we adopt Normalized Symmetry Geometric Distance (NSGD)~\cite{bian2019evaluation} to evaluate the accuracy of the estimated fundamental matrix, which is defined as the ratio of SGD~\cite{zhang1998determining} to the length of the image diagonal. Importantly, we detect up to 4$k$ keypoints for each image, and determine whether an estimate is accurate by a threshold (0.05 is used by default) for the Normalized SGD error. Following FM-Bench paper~\cite{bian2019evaluation}, we present: 1) recall (\%\textit{Recall}), the ratio of accurate estimates to all estimates evaluated by NSGD, and 2) mean number of correct correspondences (\textit{\#Corrs(-m)}) after/before RANSAC. It should be mentioned that \textit{\#Corrs(-m)} is only used for result analysis rather than performance comparison, since the influence of correspondence numbers on high-level visual applications like SfM~\cite{schonberger2016structure} is debatable~\cite{schonberger2017comparative}. However, too few matches would degrade these applications. Hence, we give minimal attention to match numbers, as long as they are not excessively low.

\subsubsection{Baselines} Similar to the relative pose estimation experiment, we select hand-crafted heuristics, NN+RT~\cite{lowe2004distinctive} or MNN, learning-based outlier pruners, OANet~\cite{zhang1998determining} and ConvMatch~\cite{zhang2023convmatch}, and learning-based matchers, SuperGlue~\cite{sarlin2020superglue}, SGMNet~\cite{chen2021learning}, and ParaFormer~\cite{lu2023paraformer} as the comparative methods.

\subsubsection{Results} The quantitative results are concluded in Tab.~\ref{FM}. \%\textit{Recall} indicates the overall performance. Notably, the performance NN+RT or MNN on KITTI is similar to learning-based matchers, since this sub-dataset is only comprised of short-baseline image pairs which are easy for the fundamental matrix estimation task. Obviously, with RootSIFT features, MaKeGNN is capable of achieving the state-of-the-art performance on all sub-datasets. With SuperPoint features, our network performs best on the first three sub-datasets (\emph{i.e.}, CPC, KITTI, and TUM) and ranks the second on T\&T.

To sum up, with well-distributed matchable keypoints as bottlenecks to guide compact and robust message passing, the proposed method yields promising results, both in short-baseline (TUM and KITTI) or wide-baseline (CPC and T\&T) scenarios.

\begin{table*}[!t]
	\caption{Evaluation on FM-Bench for fundamental matrix estimation. \%\textit{Recall} and \textit{\#Corrs(-m)} after/before RANSAC are reported. \textcolor[RGB]{254,67,101}{Red} indicates the best, \textcolor[RGB]{0,124,62}{green} ranks the second, and \textcolor[RGB]{0,90,171}{blue} ranks the third. Notably, \%\textit{Recall} indicates the overall performance.}
	\label{FM}
	\centering
	\resizebox{\textwidth}{!}{%
		\begin{tabular}{c|c|cc|cc|cc|cc}
			\toprule
			\multirow{2}{*}{Feature}    & \multirow{2}{*}{Matcher}      & \multicolumn{2}{c|}{CPC}       & \multicolumn{2}{c|}{KITTI}      & \multicolumn{2}{c|}{TUM}       & \multicolumn{2}{c}{T\&T}       \\ \cline{3-10}
			&                                        & \textit{\%Recall} & \textit{\#Corrs(-m)} & \textit{\%Recall} & \textit{\#Corrs(-m)} & \textit{\%Recall} & \textit{\#Corrs(-m)} & \textit{\%Recall} & \textit{\#Corrs(-m)} \\ \hline
			\multirow{7}{*}{RootSIFT}  & NN+RT                                  & 54.6              & 110(199)              & 90.3              & 866(995)             & 62.0              & 454(680)             & 82.1              & 246(426)             \\
			& OANet                                  & 58.6              & 119(167)             & 89.0              & 773(854)             & 62.3              & 454(396)             & 84.7              & 219(306)             \\
			& ConvMatch                                  & 52.8             & 139(237)             & 85.8              & 504(587)             & 64.4              & 576(807)             & 73.5              & 228(376)             \\
			& SuperGlue                                 & 59.0             & \textcolor[RGB]{0,90,171}{254}(\textcolor[RGB]{0,90,171}{660})             & 90.4     & \textcolor[RGB]{254,67,101}{1297}(\textcolor[RGB]{254,67,101}{2020})           & \textcolor[RGB]{0,90,171}{65.2}     & \textcolor[RGB]{0,90,171}{871}(\textcolor[RGB]{0,124,62}{1689})            & \textcolor[RGB]{0,90,171}{86.5}              & \textcolor[RGB]{0,124,62}{451}(\textcolor[RGB]{254,67,101}{1032})           \\
			& SGMNet                                 & \textcolor[RGB]{0,124,62}{61.6}             & \textcolor[RGB]{254,67,101}{284}(\textcolor[RGB]{0,124,62}{664})             & \textcolor[RGB]{0,90,171}{91.1}     & 1122(1671)           & \textcolor[RGB]{0,124,62}{66.5}     & \textcolor[RGB]{254,67,101}{942}(\textcolor[RGB]{0,90,171}{1651})          & 85.6              & \textcolor[RGB]{254,67,101}{467}(\textcolor[RGB]{0,90,171}{947})            \\
			& ParaFormer  & \textcolor[RGB]{0,90,171}{59.4}              &244(\textcolor[RGB]{254,67,101}{675})            &\textcolor[RGB]{0,124,62}{91.2}    &\textcolor[RGB]{0,124,62}{1264}(\textcolor[RGB]{0,124,62}{1883})          &62.0     &838(\textcolor[RGB]{254,67,101}{1750})           &\textcolor[RGB]{0,124,62}{87.1}              & 429(\textcolor[RGB]{0,124,62}{969})            \\
			& MaKeGNN & \textcolor[RGB]{254,67,101}{61.7}             & \textcolor[RGB]{0,124,62}{262}(658)    & \textcolor[RGB]{254,67,101}{91.6}           & \textcolor[RGB]{0,90,171}{1249}(\textcolor[RGB]{0,90,171}{1777})  & \textcolor[RGB]{254,67,101}{66.7}              & \textcolor[RGB]{0,124,62}{882}(1515)   &  \textcolor[RGB]{254,67,101}{87.4}   & \textcolor[RGB]{0,90,171}{443}(915)   \\
			\hline
			\multirow{6}{*}{SuperPoint} & MNN                                    & 34.5              & 152(421)             & \textcolor[RGB]{0,124,62}{88.6}     & 848(1490)            & 56.7              & 280(420)             & 72.8              & 287(717)             \\
			& OANet                                  & 62.9              & 186(343)             & 82.2              & 482(736)             & 61.4              & 332(473)             & 91.2              & 280(477)             \\
			& ConvMatch                                  & 55.5             & 242(942)             & 85.8             & 848(2061)             & 59.9              & 743(1996)             & 85.8              & 394(\textcolor[RGB]{0,124,62}{1337})             \\
			& SuperGlue                                 & 68.0     & 320(\textcolor[RGB]{0,90,171}{991})             & 88.1              & 956(\textcolor[RGB]{0,90,171}{2101})            & 58.7     & 725(\textcolor[RGB]{0,124,62}{2099})            & \textcolor[RGB]{0,90,171}{93.7}              & 456(1305)            \\
			& SGMNet                                 & \textcolor[RGB]{0,90,171}{69.3}     & \textcolor[RGB]{254,67,101}{377}(987)             & 88.0              & \textcolor[RGB]{0,90,171}{981}(\textcolor[RGB]{0,124,62}{2157})           & \textcolor[RGB]{254,67,101}{64.7}    & \textcolor[RGB]{254,67,101}{924}(\textcolor[RGB]{0,90,171}{2067})            & 93.6              & \textcolor[RGB]{254,67,101}{506}(1260)            \\
			& ParaFormer                                 &\textcolor[RGB]{0,124,62}{71.0}     &\textcolor[RGB]{0,90,171}{351}(\textcolor[RGB]{254,67,101}{1058})              &\textcolor[RGB]{0,90,171}{88.4}               &\textcolor[RGB]{254,67,101}{984}(\textcolor[RGB]{254,67,101}{2178})            &\textcolor[RGB]{0,90,171}{61.5}     & \textcolor[RGB]{0,124,62}{801}(\textcolor[RGB]{254,67,101}{2103})          & \textcolor[RGB]{254,67,101}{95.6}           & \textcolor[RGB]{0,90,171}{484}(\textcolor[RGB]{254,67,101}{1372})         \\
			& MaKeGNN& \textcolor[RGB]{254,67,101}{72.4}              & \textcolor[RGB]{0,124,62}{364}(\textcolor[RGB]{0,124,62}{1021})   & \textcolor[RGB]{254,67,101}{88.8}             & \textcolor[RGB]{254,67,101}{984}(2035)   & \textcolor[RGB]{254,67,101}{64.7}              & \textcolor[RGB]{0,90,171}{781}(1749)   & \textcolor[RGB]{0,124,62}{94.2}     & \textcolor[RGB]{0,124,62}{498}(\textcolor[RGB]{0,90,171}{1319})  \\
			\bottomrule
		\end{tabular}%
	}
\end{table*}

\subsection{Visual Localization}
Recent advances in feature matching learning are also conducive to visual localization, which aims to estimate the 6 degree-of-freedom (6-DOF) camera pose of a query image in relation to the corresponding 3D scene model. To be specific, visual localization remains highly challenging particularly for large-scale environments and in presence of significant illumination variation and extreme viewpoint change. Consequently, we integrate MaKeGNN into the official HLoc~\cite{sarlin2019coarse} pipeline to investigate how it benefits visual localization.

\subsubsection{Dataset} We adopt Aachen Day-Night dataset~\cite{sattler2018benchmarking} to evaluate the effectiveness of our network on visual localization. Specifically speaking, Aachen Day-Night comprises of 4328 reference images from a European old town and 922 query (824 daytime, 98 nighttime) images taken by mobile phone cameras.

\subsubsection{Evaluation Protocols} We resort to the official pipeline of Aachen Day-Night benchmark. Specifically, correspondences between reference images are first used to triangulate a 3D scene model. Then, correspondences between each query image and its retrieved reference images are established by feature matching methods to recover the relative pose. Importantly, we detect up to 8$k$ keypoints with RootSIFT and 4$k$ keypoints with SuperPoint for each image. In line with the official benchmark, we report the percentage of correctly localized queries at different distance and orientation thresholds, \emph{i.e.}, \textit{High-precision} (0.25m, 2$^\circ$), \textit{Medium-precision} (0.5m, 5$^\circ$), and \textit{Coarse-precision} (5.0m, 10$^\circ$), to reflect the visual localization performance.

\subsubsection{Baselines} We compare MakeGNN with MNN, SuperGlue~\cite{sarlin2020superglue}, SGMNet~\cite{chen2021learning}, and ParaFormer~\cite{lu2023paraformer}. In particular, we perform 50 Sinkhorn iterations for all GNN-based competitors to ensure efficiency.

\subsubsection{Results} Tab.~\ref{viusal_localization} presents the all-sided quantitative evaluation on Aachen Day-Night. Despite the larger viewpoint and illumination variations typically encountered in the long-term large-scale localization task, leading to more non-repeatable keypoints in both query and reference images, MaKeGNN consistently outperforms other baselines when using both RootSIFT and SuperPoint features. This is attributed to MaKeGNN's dynamic sampling of two sets of well-distributed matchable keypoints as bottlenecks, which avoids the interference of non-repeatable keypoints and facilitates context aggregation through a compact and robust attention pattern.

\begin{table}[t]
	\caption{Evaluation on Aachen Day-Night for visual localization. The percentage of correctly localized queries at different thresholds is reported. \textcolor[RGB]{254,67,101}{Red} indicates the best, \textcolor[RGB]{0,124,62}{green} ranks the second, and \textcolor[RGB]{0,90,171}{blue} ranks the third.}\label{viusal_localization}
	\centering
	\begin{tabular}{cccc}
		\toprule
		\multirow{2}{*}{Feature}&\multirow{2}{*}{Matcher} & \multicolumn{1}{c}{Day}                & Night              \\ \cmidrule{3-4}
		&& \multicolumn{2}{c}{(0.25m, 2°) / (0.5m, 5°) / (5.0m, 10°)}    \\
		\midrule
		\multirow{5}{*}{RootSIFT}&
		MNN& \multicolumn{1}{c}{82.3/89.1/92.1} & 43.9/56.1/65.3 \\
		&SuperGlue& \multicolumn{1}{c}{\textcolor[RGB]{0,90,171}{83.1}/\textcolor[RGB]{0,90,171}{90.9}/95.6} & \textcolor[RGB]{0,124,62}{71.4}/\textcolor[RGB]{0,124,62}{75.5}/85.7 \\
		&SGMNet&\multicolumn{1}{c}{81.8/90.2/\textcolor[RGB]{0,124,62}{96.1}} & 67.3/\textcolor[RGB]{0,124,62}{75.5}/\textcolor[RGB]{0,124,62}{93.9}\\
		&ParaFormer& \multicolumn{1}{c}{\textcolor[RGB]{0,124,62}{83.4}/\textcolor[RGB]{0,124,62}{91.1}/\textcolor[RGB]{0,90,171}{95.8}} & \textcolor[RGB]{0,90,171}{69.4}/\textcolor[RGB]{0,124,62}{75.5}/\textcolor[RGB]{0,90,171}{87.8} \\
		&MaKeGNN&\multicolumn{1}{c}{\textcolor[RGB]{254,67,101}{83.9}/\textcolor[RGB]{254,67,101}{91.3}/\textcolor[RGB]{254,67,101}{96.9}} &\textcolor[RGB]{254,67,101}{73.2}/\textcolor[RGB]{254,67,101}{79.6}/\textcolor[RGB]{254,67,101}{95.9} \\
		\midrule
		\multirow{6}{*}{SuperPoint}&
		MNN& \multicolumn{1}{c}{85.4/92.0/95.5} & 75.5/86.7/92.9 \\
		&SuperGlue& \multicolumn{1}{c}{\textcolor[RGB]{0,90,171}{86.7}/\textcolor[RGB]{0,90,171}{93.1}/\textcolor[RGB]{0,90,171}{97.1}} & 77.6/88.8/\textcolor[RGB]{0,124,62}{98.0} \\
		&SGMNet&\multicolumn{1}{c}{\textcolor[RGB]{0,124,62}{86.9}/\textcolor[RGB]{0,124,62}{93.6}/\textcolor[RGB]{0,90,171}{97.1}} & \textcolor[RGB]{0,90,171}{79.6}/\textcolor[RGB]{254,67,101}{92.9}/\textcolor[RGB]{254,67,101}{99.0}\\
		&ParaFormer& \multicolumn{1}{c}{\textcolor[RGB]{0,90,171}{86.7}/92.4/\textcolor[RGB]{254,67,101}{97.2}} & \textcolor[RGB]{0,124,62}{81.6}/\textcolor[RGB]{0,90,171}{90.8}/\textcolor[RGB]{0,124,62}{98.0}\\
		&MaKeGNN&\multicolumn{1}{c}{\textcolor[RGB]{254,67,101}{87.5}/\textcolor[RGB]{254,67,101}{93.7}/\textcolor[RGB]{254,67,101}{97.2}}  &\textcolor[RGB]{254,67,101}{82.6}/\textcolor[RGB]{254,67,101}{92.9}/\textcolor[RGB]{0,124,62}{98.0} \\
		\bottomrule
	\end{tabular}
\end{table}

\subsection{Understanding MaKeGNN}
\subsubsection{Efficiency} 
\begin{figure}[t]
	\centering
	\includegraphics[width=\linewidth]{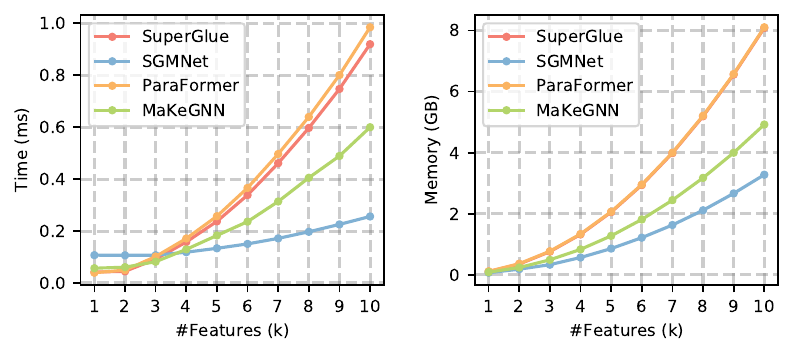}
	\caption{Computation and memory efficiency of learning-based matchers on a single GeForce RTX 3090 GPU. Notably, the memory occupation of SuperGlue is close to that of ParaFormer.}
	\label{efficiency}
\end{figure}

We compare the computation and memory efficiency of our network with SuperGlue~\cite{sarlin2020superglue}, SGMNet~\cite{chen2021learning}, and ParaFormer~\cite{lu2023paraformer}. Fig.~\ref{efficiency} summarizes the statistical results \emph{w.r.t.} the number of input 128-D features, using a single GeForce RTX 3090 GPU with 24GB memory. In particular, memory occupation are averaged by batch size during the training process. Apparently, since MaKeGNN leverages well-distributed matchable keypoints as bottlenecks to guide compact message passing, it reduces the runtime and memory occupation on a large number of keypoints (10$k$) by 34.77\% and 39.14\%, respectively, compared to SuperGlue. Although SGMNet exhibits greater efficiency in terms of time and memory, the proposed method deliver significantly better performance while maintaining modest time and memory efficiency. In other words, MaKeGNN achieves an improved trade-off between accuracy and efficiency, rendering it more suitable for demanding real-time visual tasks.

\subsubsection{Impact of Sampled Matchable Keypoint Number} At the heart of MaKeGNN lies the dynamic sampling of two sets of well-distributed keypoints with high matchability scores, which serve as message bottlenecks for compact and robust message passing. Consequently, the number of sampled matchable keypoints has a direct impact on both the efficiency and accuracy of MaKeGNN. To
investigate this impact, we further perform a grid search for different keypoint and sampling numbers, on ScanNet dataset using RootSIFT features.

\begin{figure}[t]
	\centering
	\includegraphics[width=0.58\linewidth]{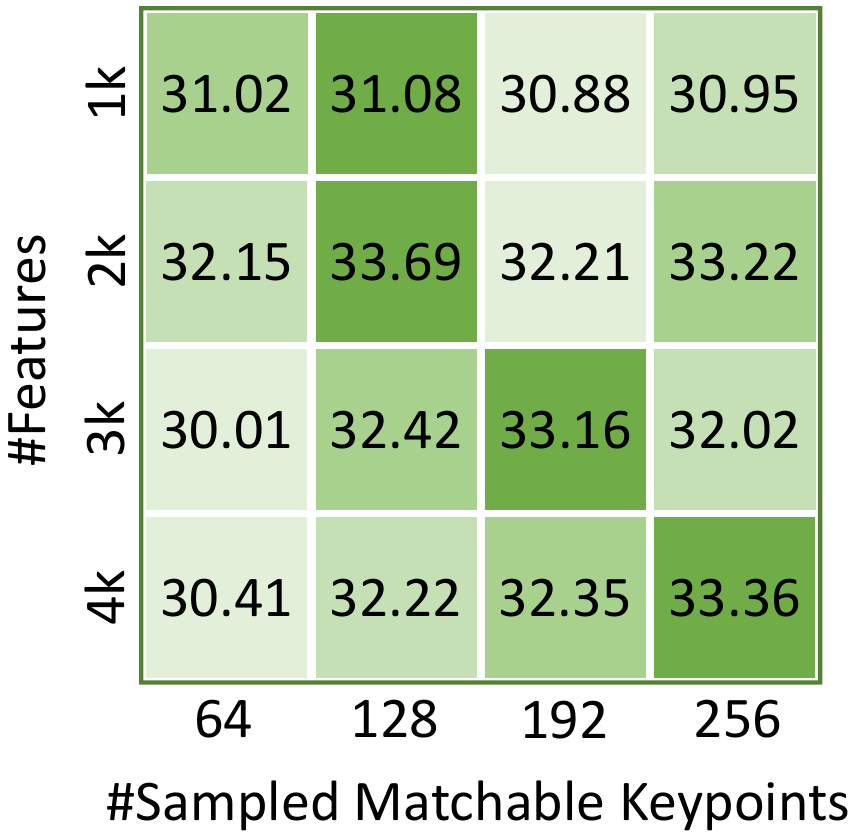}
	\caption{The impact of the sampled matchable keypoint number when varying the input keypoint number. Numbers in grids are AUC@$5^{\circ}$ on ScanNet using RootSIFT features.}
	\label{seed_num}
\end{figure}

As depicted in Fig.~\ref{seed_num}, maintaining an approximate proportional relationship between the number of input keypoints and sampled matchable ones delivers the optimal performance. The reason for this is that sampling an excessive number of matchable keypoints may introduce several insignificant message interaction, resulting in less reliable guidance. Conversely, sampling too few matchable keypoints can result in significant information loss.

\subsubsection{Ablation Study} To investigate the contribution of each design presented in our approach, we conduct a comprehensive variety of ablation studies on ScanNet dataset using SuperPoint, and the results are concluded in Tab.~\ref{ablation_study}. Row 6 shows our complete model as a point of reference.
Row 3 reveals that compared to only leveraging per-keypoint feature for matchabilty score prediction, combining it with bilateral context is a better choice, since this combination results in a significant improvement in the precision and recall of the sampled matchable keypoints, as indicated in Tab.~\ref{predictor_PR}. Row 4 validates that compared to the vanilla attention, our matchability score-guided attention that is tailor-made for the feature matching task, can suppress the interference of non-repeatable keypoints caused by occlusion and failure of the detector, empowering the network to achieve cleaner and sharper message passing. Row 5 further proves that sampling well-distributed matchable keypoints as message bottlenecks plays an important role, which is capable of guiding information interaction across images for robust feature matching.

In summary, each component in our network contributes to the final performance notably.

\begin{table}[!t]
	\centering
		\caption{Ablation study of MaKeGNN. w/o. B.C. stands for only using per-keypoint feature instead of concatenating it with Bilateral Context (\emph{i.e.}, global representation vectors $\mathbf{f}_\mathbf{A}^g$ and $\mathbf{f}_\mathbf{B}^g$), to predict the matchability score. w/o. M.G.A.A. stands for replacing the matchability score-guided attention with the vanilla attention for context aggregation. w. Rand. Sampling means that sampling keypoints randomly instead of pick the top-\textit{k} matchability scores processed by NMS. AUC@$20^{\circ}$, M.S., Prec. on ScanNet using SuperPoint features are reported. \textbf{Bold} indicates the best.}
	\label{ablation_study}
	\resizebox{\linewidth}{!}{
		\begin{tabular}{lccc}
			\toprule
			Matcher&AUC@$20^{\circ}$&M.S. & Prec.  \\
			\midrule
			(1) MNN&38.00&8.72&44.54 \\
			(2) SuperGlue&54.89&16.25&46.16 \\
			\midrule
			(3) MaKeGNN w/o. B.C.&56.43&16.50&48.12 \\
			(4) MaKeGNN w/o. M.G.A.A.&57.12&16.61&48.33 \\
			(5) MaKeGNN w. Rand. Sampling&51.24&15.33&39.80\\
			\midrule
			(6) MaKeGNN full&\textbf{58.04}&\textbf{16.88}&\textbf{48.39} \\
			\bottomrule
	\end{tabular}}
\end{table}

\begin{table}[!t]
	\centering
	\caption{Precision and recall of matchable keypoints sampled by the last BCAS module. A keypoint is considered matchable, if its matchability score larger than 0.5.}
	\label{predictor_PR}
	\begin{tabular}{lcc}
			\toprule
			Sampler&Precision&Recall  \\
			\midrule
			BCAS w/o. B.C.&85.63&91.12\\
			BCAS&96.87&96.19\\
			\bottomrule
	\end{tabular}
\end{table}

\subsubsection{Visualization of Sampled Matchable Keypoints} 
\begin{figure}[t]
	\centering
	\includegraphics[width=\linewidth]{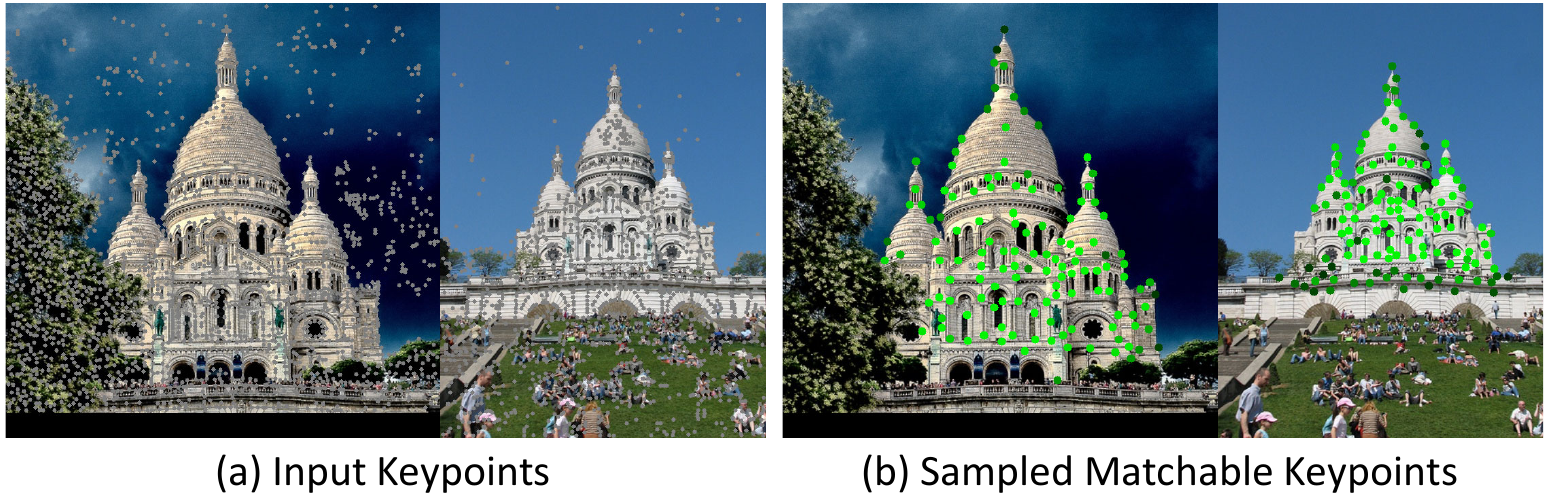}
	\caption{Example of (a) input keypoints and (b) sampled matchable keypoints.}
	\label{visualization}
\end{figure}

We visualize the sampled matchable keypoints in the image pair, using the proposed BCAS module, as depicted in Fig.~\ref{visualization}. The brightness of green corresponds to the matchability score, with brighter green indicating higher scores. Obviously, the sampled keypoints with high matchability scores span almost the whole meaningful regions and are positioned in significant structures, thereby providing fruitful geometric and visual cues to facilitate the context aggregation process.

%
%
%
%
\section{Conclusion}
This paper presents MaKeGNN, a top-performing graph neural network for feature matching learning. With the proposed \emph{Bilateral Context-Aware Sampling} (BCAS) and \emph{Matchable Keypoint-Assisted Context Aggregation} (MKACA) modules, MaKeGNN is capable of aggregating context with a compact and robust attention pattern. Comprehensive experiments on diverse tasks and datasets demonstrate that our method outperforms the strongest prior model with modest computation and memory costs.

\bibliographystyle{IEEEtran}
\bibliography{references}

\begin{thebibliography}{10}
\providecommand{\url}[1]{#1}
\csname url@samestyle\endcsname
\providecommand{\newblock}{\relax}
\providecommand{\bibinfo}[2]{#2}
\providecommand{\BIBentrySTDinterwordspacing}{\spaceskip=0pt\relax}
\providecommand{\BIBentryALTinterwordstretchfactor}{4}
\providecommand{\BIBentryALTinterwordspacing}{\spaceskip=\fontdimen2\font plus
\BIBentryALTinterwordstretchfactor\fontdimen3\font minus
  \fontdimen4\font\relax}
\providecommand{\BIBforeignlanguage}[2]{{%
\expandafter\ifx\csname l@#1\endcsname\relax
\typeout{** WARNING: IEEEtran.bst: No hyphenation pattern has been}%
\typeout{** loaded for the language `#1'. Using the pattern for}%
\typeout{** the default language instead.}%
\else
\language=\csname l@#1\endcsname
\fi
#2}}
\providecommand{\BIBdecl}{\relax}
\BIBdecl

\bibitem{ma2021image}
J.~Ma, X.~Jiang, A.~Fan, J.~Jiang, and J.~Yan, ``Image matching from
  handcrafted to deep features: A survey,'' \emph{International Journal of
  Computer Vision}, vol. 129, pp. 23--79, 2021.

\bibitem{mur2015orb}
R.~Mur-Artal, J.~M.~M. Montiel, and J.~D. Tardos, ``Orb-slam: a versatile and
  accurate monocular slam system,'' \emph{IEEE Transactions on Robotics},
  vol.~31, no.~5, pp. 1147--1163, 2015.

\bibitem{schonberger2016structure}
J.~L. Schonberger and J.-M. Frahm, ``Structure-from-motion revisited,'' in
  \emph{Proceedings of the IEEE Conference on Computer Vision and Pattern
  Recognition}, 2016, pp. 4104--4113.

\bibitem{sarlin2019coarse}
P.-E. Sarlin, C.~Cadena, R.~Siegwart, and M.~Dymczyk, ``From coarse to fine:
  Robust hierarchical localization at large scale,'' in \emph{Proceedings of
  the IEEE Conference on Computer Vision and Pattern Recognition}, 2019, pp.
  12\,716--12\,725.

\bibitem{zhang2021image}
H.~Zhang, H.~Xu, X.~Tian, J.~Jiang, and J.~Ma, ``Image fusion meets deep
  learning: A survey and perspective,'' \emph{Information Fusion}, vol.~76, pp.
  323--336, 2021.

\bibitem{sarlin2020superglue}
P.-E. Sarlin, D.~DeTone, T.~Malisiewicz, and A.~Rabinovich, ``Superglue:
  Learning feature matching with graph neural networks,'' in \emph{Proceedings
  of the IEEE Conference on Computer Vision and Pattern Recognition}, 2020, pp.
  4938--4947.

\bibitem{lu2023robust}
Y.~Lu, J.~Ma, L.~Fang, X.~Tian, and J.~Jiang, ``Robust and scalable gaussian
  process regression and its applications,'' in \emph{Proceedings of the IEEE
  Conference on Computer Vision and Pattern Recognition}, 2023, pp.
  21\,950--21\,959.

\bibitem{yi2018learning}
K.~M. Yi, E.~Trulls, Y.~Ono, V.~Lepetit, M.~Salzmann, and P.~Fua, ``Learning to
  find good correspondences,'' in \emph{Proceedings of the IEEE Conference on
  Computer Vision and Pattern Recognition}, 2018, pp. 2666--2674.

\bibitem{zhang2019learning}
J.~Zhang, D.~Sun, Z.~Luo, A.~Yao, L.~Zhou, T.~Shen, Y.~Chen, L.~Quan, and
  H.~Liao, ``Learning two-view correspondences and geometry using order-aware
  network,'' in \emph{Proceedings of the IEEE International Conference on
  Computer Vision}, 2019, pp. 5845--5854.

\bibitem{sun2020acne}
W.~Sun, W.~Jiang, E.~Trulls, A.~Tagliasacchi, and K.~M. Yi, ``Acne: Attentive
  context normalization for robust permutation-equivariant learning,'' in
  \emph{Proceedings of the IEEE Conference on Computer Vision and Pattern
  Recognition}, 2020, pp. 11\,286--11\,295.

\bibitem{zhao2021progressive}
C.~Zhao, Y.~Ge, F.~Zhu, R.~Zhao, H.~Li, and M.~Salzmann, ``Progressive
  correspondence pruning by consensus learning,'' in \emph{Proceedings of the
  IEEE International Conference on Computer Vision}, 2021, pp. 6464--6473.

\bibitem{liu2021learnable}
Y.~Liu, L.~Liu, C.~Lin, Z.~Dong, and W.~Wang, ``Learnable motion coherence for
  correspondence pruning,'' in \emph{Proceedings of the IEEE Conference on
  Computer Vision and Pattern Recognition}, 2021, pp. 3237--3246.

\bibitem{dai2022ms2dg}
L.~Dai, Y.~Liu, J.~Ma, L.~Wei, T.~Lai, C.~Yang, and R.~Chen, ``Ms2dg-net:
  Progressive correspondence learning via multiple sparse semantics dynamic
  graph,'' in \emph{Proceedings of the IEEE Conference on Computer Vision and
  Pattern Recognition}, 2022, pp. 8973--8982.

\bibitem{zheng2022msa}
L.~Zheng, G.~Xiao, Z.~Shi, S.~Wang, and J.~Ma, ``Msa-net: Establishing reliable
  correspondences by multiscale attention network,'' \emph{IEEE Transactions on
  Image Processing}, vol.~31, pp. 4598--4608, 2022.

\bibitem{zhang2023convmatch}
S.~Zhang and J.~Ma, ``Convmatch: Rethinking network design for two-view
  correspondence learning,'' in \emph{Proceedings of the AAAI Conference on
  Artificial Intelligence}, 2023.

\bibitem{li2023umatch}
Z.~Li, S.~Zhang, and J.~Ma, ``U-match: Two-view correspondence learning with
  hierarchy-aware local context aggregation,'' in \emph{Proceedings of the
  International Joint Conference on Artificial Intelligence}, 2023.

\bibitem{li2023two}
Z.~Li, Y.~Ma, X.~Mei, and J.~Ma, ``Two-view correspondence learning using graph
  neural network with reciprocal neighbor attention,'' \emph{ISPRS Journal of
  Photogrammetry and Remote Sensing}, vol. 202, pp. 114--124, 2023.

\bibitem{liu2023pgfnet}
X.~Liu, G.~Xiao, R.~Chen, and J.~Ma, ``Pgfnet: Preference-guided filtering
  network for two-view correspondence learning,'' \emph{IEEE Transactions on
  Image Processing}, vol.~32, pp. 1367--1378, 2023.

\bibitem{qi2017pointnet}
C.~R. Qi, H.~Su, K.~Mo, and L.~J. Guibas, ``Pointnet: Deep learning on point
  sets for 3d classification and segmentation,'' in \emph{Proceedings of the
  IEEE Conference on Computer Vision and Pattern Recognition}, 2017, pp.
  652--660.

\bibitem{suwanwimolkul2022efficient}
S.~Suwanwimolkul and S.~Komorita, ``Efficient linear attention for fast and
  accurate keypoint matching,'' in \emph{Proceedings of the International
  Conference on Multimedia Retrieval}, 2022, pp. 330--341.

\bibitem{cai2023htmatch}
Y.~Cai, L.~Li, D.~Wang, X.~Li, and X.~Liu, ``Htmatch: An efficient hybrid
  transformer based graph neural network for local feature matching,''
  \emph{Signal Processing}, vol. 204, p. 108859, 2023.

\bibitem{thomee2016yfcc100m}
B.~Thomee, D.~A. Shamma, G.~Friedland, B.~Elizalde, K.~Ni, D.~Poland, D.~Borth,
  and L.-J. Li, ``Yfcc100m: The new data in multimedia research,''
  \emph{Communications of the ACM}, vol.~59, no.~2, pp. 64--73, 2016.

\bibitem{chen2021learning}
H.~Chen, Z.~Luo, J.~Zhang, L.~Zhou, X.~Bai, Z.~Hu, C.-L. Tai, and L.~Quan,
  ``Learning to match features with seeded graph matching network,'' in
  \emph{Proceedings of the IEEE International Conference on Computer Vision},
  2021, pp. 6301--6310.

\bibitem{shi2022clustergnn}
Y.~Shi, J.-X. Cai, Y.~Shavit, T.-J. Mu, W.~Feng, and K.~Zhang, ``Clustergnn:
  Cluster-based coarse-to-fine graph neural network for efficient feature
  matching,'' in \emph{Proceedings of the IEEE Conference on Computer Vision
  and Pattern Recognition}, 2022, pp. 12\,517--12\,526.

\bibitem{lowe2004distinctive}
D.~G. Lowe, ``Distinctive image features from scale-invariant keypoints,''
  \emph{International Journal of Computer Vision}, vol.~60, pp. 91--110, 2004.

\bibitem{rublee2011orb}
E.~Rublee, V.~Rabaud, K.~Konolige, and G.~Bradski, ``Orb: An efficient
  alternative to sift or surf,'' in \emph{Proceedings of the IEEE International
  Conference on Computer Vision}.\hskip 1em plus 0.5em minus 0.4em\relax Ieee,
  2011, pp. 2564--2571.

\bibitem{detone2018superpoint}
D.~DeTone, T.~Malisiewicz, and A.~Rabinovich, ``Superpoint: Self-supervised
  interest point detection and description,'' in \emph{Proceedings of the IEEE
  Conference on Computer Vision and Pattern Recognition}, 2018, pp. 224--236.

\bibitem{dusmanu2019d2}
M.~Dusmanu, I.~Rocco, T.~Pajdla, M.~Pollefeys, J.~Sivic, A.~Torii, and
  T.~Sattler, ``D2-net: A trainable cnn for joint description and detection of
  local features,'' in \emph{Proceedings of the IEEE Conference on Computer
  Vision and Pattern Recognition}, 2019, pp. 8092--8101.

\bibitem{revaud2019r2d2}
J.~Revaud, C.~De~Souza, M.~Humenberger, and P.~Weinzaepfel, ``R2d2: Reliable
  and repeatable detector and descriptor,'' in \emph{Proceedings of the
  Advances in Neural Information Processing Systems}, vol.~32, 2019.

\bibitem{wang2023attention}
C.~Wang, R.~Xu, K.~Lv, S.~Xu, W.~Meng, Y.~Zhang, B.~Fan, and X.~Zhang,
  ``Attention weighted local descriptors,'' \emph{IEEE Transactions on Pattern
  Analysis and Machine Intelligence}, 2023.

\bibitem{lu2023paraformer}
X.~Lu, Y.~Yan, B.~Kang, and S.~Du, ``Paraformer: Parallel attention transformer
  for efficient feature matching,'' in \emph{Proceedings of the AAAI Conference
  on Artificial Intelligence}, 2023.

\bibitem{li2022guided}
Z.~Li, Y.~Ma, X.~Mei, J.~Huang, and J.~Ma, ``Guided neighborhood affine
  subspace embedding for feature matching,'' \emph{Pattern Recognition}, vol.
  124, p. 108489, 2022.

\bibitem{ying2018hierarchical}
Z.~Ying, J.~You, C.~Morris, X.~Ren, W.~Hamilton, and J.~Leskovec,
  ``Hierarchical graph representation learning with differentiable pooling,''
  in \emph{Proceedings of the Advances in Neural Information Processing
  Systems}, vol.~31, 2018.

\bibitem{vaswani2017attention}
A.~Vaswani, N.~Shazeer, N.~Parmar, J.~Uszkoreit, L.~Jones, A.~N. Gomez,
  {\L}.~Kaiser, and I.~Polosukhin, ``Attention is all you need,'' in
  \emph{Proceedings of the Advances in Neural Information Processing Systems},
  vol.~30, 2017.

\bibitem{wang2019learning}
R.~Wang, J.~Yan, and X.~Yang, ``Learning combinatorial embedding networks for
  deep graph matching,'' in \emph{Proceedings of the IEEE International
  Conference on Computer Vision}, 2019, pp. 3056--3065.

\bibitem{dosovitskiy2020image}
A.~Dosovitskiy, L.~Beyer, A.~Kolesnikov, D.~Weissenborn, X.~Zhai,
  T.~Unterthiner, M.~Dehghani, M.~Minderer, G.~Heigold, S.~Gelly \emph{et~al.},
  ``An image is worth 16x16 words: Transformers for image recognition at
  scale,'' in \emph{Proceedings of the International Conference on Learning
  Representations}, 2020.

\bibitem{carion2020end}
N.~Carion, F.~Massa, G.~Synnaeve, N.~Usunier, A.~Kirillov, and S.~Zagoruyko,
  ``End-to-end object detection with transformers,'' in \emph{Proceedings of
  the European Conference on Computer Vision}, 2020, pp. 213--229.

\bibitem{wang2020axial}
H.~Wang, Y.~Zhu, B.~Green, H.~Adam, A.~Yuille, and L.-C. Chen, ``Axial-deeplab:
  Stand-alone axial-attention for panoptic segmentation,'' in \emph{Proceedings
  of the European Conference on Computer Vision}, 2020, pp. 108--126.

\bibitem{sun2021loftr}
J.~Sun, Z.~Shen, Y.~Wang, H.~Bao, and X.~Zhou, ``Loftr: Detector-free local
  feature matching with transformers,'' in \emph{Proceedings of the IEEE
  Conference on Computer Vision and Pattern Recognition}, 2021, pp. 8922--8931.

\bibitem{kitaev2020reformer}
N.~Kitaev, {\L}.~Kaiser, and A.~Levskaya, ``Reformer: The efficient
  transformer,'' in \emph{Proceedings of the International Conference on
  Learning Representations}, 2020.

\bibitem{tay2020sparse}
Y.~Tay, D.~Bahri, L.~Yang, D.~Metzler, and D.-C. Juan, ``Sparse sinkhorn
  attention,'' in \emph{Proceedings of the International Conference on Machine
  Learning}, 2020, pp. 9438--9447.

\bibitem{katharopoulos2020transformers}
A.~Katharopoulos, A.~Vyas, N.~Pappas, and F.~Fleuret, ``Transformers are rnns:
  Fast autoregressive transformers with linear attention,'' in
  \emph{Proceedings of the International Conference on Machine Learning}, 2020,
  pp. 5156--5165.

\bibitem{wang2020linformer}
S.~Wang, B.~Z. Li, M.~Khabsa, H.~Fang, and H.~Ma, ``Linformer: Self-attention
  with linear complexity,'' \emph{arXiv preprint arXiv:2006.04768}, 2020.

\bibitem{chen2021psvit}
B.~Chen, P.~Li, B.~Li, C.~Li, L.~Bai, C.~Lin, M.~Sun, J.~Yan, and W.~Ouyang,
  ``Psvit: Better vision transformer via token pooling and attention sharing,''
  \emph{arXiv preprint arXiv:2108.03428}, 2021.

\bibitem{tay2022efficient}
Y.~Tay, M.~Dehghani, D.~Bahri, and D.~Metzler, ``Efficient transformers: A
  survey,'' \emph{ACM Computing Surveys}, vol.~55, no.~6, pp. 1--28, 2022.

\bibitem{fischler1981random}
M.~A. Fischler and R.~C. Bolles, ``Random sample consensus: a paradigm for
  model fitting with applications to image analysis and automated
  cartography,'' \emph{Communications of the ACM}, vol.~24, no.~6, pp.
  381--395, 1981.

\bibitem{raguram2012usac}
R.~Raguram, O.~Chum, M.~Pollefeys, J.~Matas, and J.-M. Frahm, ``Usac: A
  universal framework for random sample consensus,'' \emph{IEEE Transactions on
  Pattern Analysis and Machine Intelligence}, vol.~35, no.~8, pp. 2022--2038,
  2012.

\bibitem{barath2018graph}
D.~Barath and J.~Matas, ``Graph-cut ransac,'' in \emph{Proceedings of the IEEE
  Conference on Computer Vision and Pattern Recognition}, 2018, pp. 6733--6741.

\bibitem{barath2020magsac++}
D.~Barath, J.~Noskova, M.~Ivashechkin, and J.~Matas, ``Magsac++, a fast,
  reliable and accurate robust estimator,'' in \emph{Proceedings of the IEEE
  Conference on Computer Vision and Pattern Recognition}, 2020, pp. 1304--1312.

\bibitem{cuturi2013sinkhorn}
M.~Cuturi, ``Sinkhorn distances: Lightspeed computation of optimal transport,''
  in \emph{Proceedings of the Advances in Neural Information Processing
  Systems}, vol.~26, 2013.

\bibitem{velivckovic2017graph}
P.~Veli{\v{c}}kovi{\'c}, G.~Cucurull, A.~Casanova, A.~Romero, P.~Lio, and
  Y.~Bengio, ``Graph attention networks,'' \emph{arXiv preprint
  arXiv:1710.10903}, 2017.

\bibitem{shen2019matchable}
T.~Shen, Z.~Luo, L.~Zhou, R.~Zhang, S.~Zhu, T.~Fang, and L.~Quan, ``Matchable
  image retrieval by learning from surface reconstruction,'' in
  \emph{Proceedings of the Asian Conference on Computer Vision}, 2019, pp.
  415--431.

\bibitem{dai2017bundlefusion}
A.~Dai, M.~Nie{\ss}ner, M.~Zollh{\"o}fer, S.~Izadi, and C.~Theobalt,
  ``Bundlefusion: Real-time globally consistent 3d reconstruction using
  on-the-fly surface reintegration,'' \emph{ACM Transactions on Graphics},
  vol.~36, no.~4, p.~1, 2017.

\bibitem{bae2022multi}
G.~Bae, I.~Budvytis, and R.~Cipolla, ``Multi-view depth estimation by fusing
  single-view depth probability with multi-view geometry,'' in
  \emph{Proceedings of the IEEE Conference on Computer Vision and Pattern
  Recognition}, 2022, pp. 2842--2851.

\bibitem{arandjelovic2012three}
R.~Arandjelovi{\'c} and A.~Zisserman, ``Three things everyone should know to
  improve object retrieval,'' in \emph{Proceedings of the IEEE Conference on
  Computer Vision and Pattern Recognition}, 2012, pp. 2911--2918.

\bibitem{cavalli2020handcrafted}
L.~Cavalli, V.~Larsson, M.~R. Oswald, T.~Sattler, and M.~Pollefeys,
  ``Handcrafted outlier detection revisited,'' in \emph{Proceedings of the
  European Conference on Computer Vision}, 2020, pp. 770--787.

\bibitem{bian2019evaluation}
J.-W. Bian, Y.-H. Wu, J.~Zhao, Y.~Liu, L.~Zhang, M.-M. Cheng, and I.~Reid, ``An
  evaluation of feature matchers for fundamental matrix estimation,'' in
  \emph{Proceedings of the British Machine Vision Conference}, 2019.

\bibitem{sturm2012benchmark}
J.~Sturm, N.~Engelhard, F.~Endres, W.~Burgard, and D.~Cremers, ``A benchmark
  for the evaluation of rgb-d slam systems,'' in \emph{Proceedings of the IEEE
  International Conference on Intelligent Robots and Systems}, 2012, pp.
  573--580.

\bibitem{geiger2012we}
A.~Geiger, P.~Lenz, and R.~Urtasun, ``Are we ready for autonomous driving? the
  kitti vision benchmark suite,'' in \emph{Proceedings of the IEEE Conference
  on Computer Vision and Pattern recognition}, 2012, pp. 3354--3361.

\bibitem{knapitsch2017tanks}
A.~Knapitsch, J.~Park, Q.-Y. Zhou, and V.~Koltun, ``Tanks and temples:
  Benchmarking large-scale scene reconstruction,'' \emph{ACM Transactions on
  Graphics}, vol.~36, no.~4, pp. 1--13, 2017.

\bibitem{wilson2014robust}
K.~Wilson and N.~Snavely, ``Robust global translations with 1dsfm,'' in
  \emph{Proceedings of the European Conference on Computer Vision}, 2014, pp.
  61--75.

\bibitem{zhang1998determining}
Z.~Zhang, ``Determining the epipolar geometry and its uncertainty: A review,''
  \emph{International Journal of Computer Vision}, vol.~27, pp. 161--195, 1998.

\bibitem{schonberger2017comparative}
J.~L. Schonberger, H.~Hardmeier, T.~Sattler, and M.~Pollefeys, ``Comparative
  evaluation of hand-crafted and learned local features,'' in \emph{Proceedings
  of the IEEE Conference on Computer Vision and Pattern Recognition}, 2017, pp.
  1482--1491.

\bibitem{sattler2018benchmarking}
T.~Sattler, W.~Maddern, C.~Toft, A.~Torii, L.~Hammarstrand, E.~Stenborg,
  D.~Safari, M.~Okutomi, M.~Pollefeys, J.~Sivic \emph{et~al.}, ``Benchmarking
  6dof outdoor visual localization in changing conditions,'' in
  \emph{Proceedings of the IEEE Conference on Computer Vision and Pattern
  Recognition}, 2018, pp. 8601--8610.

\end{thebibliography}

%

\end{document}